\documentclass[journal]{IEEEtran}

\usepackage[table]{xcolor}
\definecolor{rblue}{rgb}{0,0.5,1}
\definecolor{hollywoodcerise}{rgb}{0.96, 0.0, 0.63}
\definecolor{lasallegreen}{rgb}{0.03, 0.47, 0.19}
\definecolor{hanpurple}{rgb}{0.32, 0.09, 0.98}
\definecolor{green(pigment)}{rgb}{0.0, 0.65, 0.31}

\usepackage[pagebackref=false,breaklinks=true,colorlinks,bookmarks=false]{hyperref}
\hypersetup{colorlinks,linkcolor={red},citecolor={hanpurple},urlcolor={magenta}}  
\usepackage{url}

\definecolor{revisionred}{rgb}{0.75,0,0}

\usepackage{color}
\usepackage{xcolor}
\usepackage{booktabs}
\usepackage{colortbl}
\usepackage{amsmath}
\usepackage{amssymb}
\usepackage{multirow}
\usepackage{graphicx}
\usepackage{wrapfig}
\usepackage{float}
\usepackage{pifont}
 \usepackage{subfig}

\definecolor{mygreen}{RGB}{0, 176, 80}

\newcommand{\SecondBest}[1]{\textcolor{black}{\underline{{#1}}}}
\newcommand{\Best}[1]{\textcolor{black}{\textbf{{#1}}}}

\newcolumntype{C}[1]{>{\centering\arraybackslash}p{#1}} 

\definecolor{White}{rgb}{1.,0.,1.}
\definecolor{first}{rgb}{.8,.0,.0}
\definecolor{second}{rgb}{.0,.6,.0}
\definecolor{third}{rgb}{.0,.0,.8}

\definecolor{ceiling}{RGB}{214,  38, 40}
\definecolor{floor}{RGB}{43, 160, 4}
\definecolor{wall}{RGB}{158, 216, 229}
\definecolor{window}{RGB}{114, 158, 206}
\definecolor{chair}{RGB}{204, 204, 91}
\definecolor{bed}{RGB}{255, 186, 119}
\definecolor{sofa}{RGB}{147, 102, 188}
\definecolor{table}{RGB}{30, 119, 181}
\definecolor{tvs}{RGB}{160, 188, 33}
\definecolor{furniture}{RGB}{255, 127, 12}
\definecolor{objects}{RGB}{196, 175, 214}

\definecolor{car}{rgb}{0.39215686, 0.58823529, 0.96078431}
\definecolor{bicycle}{rgb}{0.39215686, 0.90196078, 0.96078431}
\definecolor{motorcycle}{rgb}{0.11764706, 0.23529412, 0.58823529}
\definecolor{truck}{rgb}{0.31372549, 0.11764706, 0.70588235}
\definecolor{othervehicle}{rgb}{0.39215686, 0.31372549, 0.98039216}
\definecolor{person}{rgb}{1.        , 0.11764706, 0.11764706}
\definecolor{bicyclist}{rgb}{1.        , 0.15686275, 0.78431373}
\definecolor{motorcyclist}{rgb}{0.58823529, 0.11764706, 0.35294118}
\definecolor{road}{rgb}{1.        , 0.        , 1.        }
\definecolor{parking}{rgb}{1.        , 0.58823529, 1.        }
\definecolor{sidewalk}{rgb}{0.29411765, 0.        , 0.29411765}
\definecolor{otherground}{rgb}{0.68627451, 0.        , 0.29411765}
\definecolor{building}{rgb}{1.        , 0.78431373, 0.        }
\definecolor{fence}{rgb}{1.        , 0.47058824, 0.19607843}
\definecolor{vegetation}{rgb}{0.        , 0.68627451, 0.        }
\definecolor{trunk}{rgb}{0.52941176, 0.23529412, 0.        }
\definecolor{terrain}{rgb}{0.58823529, 0.94117647, 0.31372549}
\definecolor{pole}{rgb}{1.        , 0.94117647, 0.58823529}
\definecolor{trafficsign}{rgb}{1.        , 0.        , 0.        }
\definecolor{otherstructure}{rgb}{0.98039215, 0.58823529, 0.}
\definecolor{otherobject}{rgb}{0.19607843, 1.        , 1.        }

\makeatletter
\newcommand{\car@semkitfreq}{3.92}
\newcommand{\bicycle@semkitfreq}{0.03}
\newcommand{\motorcycle@semkitfreq}{0.03}
\newcommand{\truck@semkitfreq}{0.16}
\newcommand{\othervehicle@semkitfreq}{0.20}
\newcommand{\person@semkitfreq}{0.07}
\newcommand{\bicyclist@semkitfreq}{0.07}
\newcommand{\motorcyclist@semkitfreq}{0.05}
\newcommand{\road@semkitfreq}{15.30}
\newcommand{\parking@semkitfreq}{1.12}
\newcommand{\sidewalk@semkitfreq}{11.13}
\newcommand{\otherground@semkitfreq}{0.56}
\newcommand{\building@semkitfreq}{14.1}
\newcommand{\fence@semkitfreq}{3.90}
\newcommand{\vegetation@semkitfreq}{39.3}
\newcommand{\trunk@semkitfreq}{0.51}
\newcommand{\terrain@semkitfreq}{9.17}
\newcommand{\pole@semkitfreq}{0.29}
\newcommand{\trafficsign@semkitfreq}{0.08}
\newcommand{\semkitfreq}[1]{{\csname #1@semkitfreq\endcsname}}

\newcommand{\car@sscbkitfreq}{2.85}
\newcommand{\bicycle@sscbkitfreq}{0.01}
\newcommand{\motorcycle@sscbkitfreq}{0.01}
\newcommand{\truck@sscbkitfreq}{0.16}
\newcommand{\othervehicle@sscbkitfreq}{5.75}
\newcommand{\person@sscbkitfreq}{0.02}
\newcommand{\road@sscbkitfreq}{14.98}
\newcommand{\parking@sscbkitfreq}{2.31}
\newcommand{\sidewalk@sscbkitfreq}{6.43}
\newcommand{\otherground@sscbkitfreq}{2.05}
\newcommand{\building@sscbkitfreq}{15.67}
\newcommand{\fence@sscbkitfreq}{0.96}
\newcommand{\vegetation@sscbkitfreq}{41.99}
\newcommand{\terrain@sscbkitfreq}{7.10}
\newcommand{\pole@sscbkitfreq}{0.22}
\newcommand{\trafficsign@sscbkitfreq}{0.06}
\newcommand{\otherstructure@sscbkitfreq}{4.33}
\newcommand{\otherobject@sscbkitfreq}{0.28}
\newcommand{\sscbkitfreq}[1]{{\csname #1@sscbkitfreq\endcsname}}

\usepackage{makecell}
\definecolor{colorVehicle}{RGB}{100, 150, 245}
\definecolor{colorPedestrian}{RGB}{255, 30, 30}
\definecolor{colorRoad}{RGB}{255, 0, 255}
\definecolor{colorBuilding}{RGB}{255, 200, 0}
\definecolor{colorVegetation}{RGB}{0, 175, 0}
\definecolor{colorTerrain}{RGB}{150, 240, 80}
\makeatletter
\newcommand{\vehicle@quadoccfreq}{0.90}
\newcommand{\pedestrian@quadoccfreq}{2.09} 
\newcommand{\road@quadoccfreq}{52.34}
\newcommand{\building@quadoccfreq}{14.00}
\newcommand{\vegetation@quadoccfreq}{22.82}
\newcommand{\terrain@quadoccfreq}{7.85}
\newcommand{\quadoccfreq}[1]{{\csname #1@quadoccfreq\endcsname}}
\makeatother

\definecolor{hthreeoRoad}{RGB}{128,64,128}        
\definecolor{hthreeoSidewalk}{RGB}{244,35,232}    
\definecolor{hthreeoBuilding}{RGB}{70,70,70}      
\definecolor{hthreeoVegetation}{RGB}{107,142,35}  
\definecolor{hthreeoCar}{RGB}{0,0,142}            
\definecolor{hthreeoTruck}{RGB}{0,0,70}           
\definecolor{hthreeoBus}{RGB}{0,60,100}           
\definecolor{hthreeoTwoWheeler}{RGB}{0,0,230}     
\definecolor{hthreeoPerson}{RGB}{220,20,60}       
\definecolor{hthreeoPole}{RGB}{153,153,153}       

\makeatletter
\newcommand{\road@hthreeofreqHomo}{34.04}
\newcommand{\sidewalk@hthreeofreqHomo}{26.24}
\newcommand{\building@hthreeofreqHomo}{10.00}
\newcommand{\vegetation@hthreeofreqHomo}{24.02}
\newcommand{\car@hthreeofreqHomo}{2.92}
\newcommand{\truck@hthreeofreqHomo}{0.99}
\newcommand{\bus@hthreeofreqHomo}{0.25}
\newcommand{\twoWheeler@hthreeofreqHomo}{0.07}
\newcommand{\person@hthreeofreqHomo}{0.97}
\newcommand{\pole@hthreeofreqHomo}{0.50}
\newcommand{\hthreeofreqHomo}[1]{{\csname #1@hthreeofreqHomo\endcsname}}

\newcommand{\road@hthreeofreqHeter}{34.04}
\newcommand{\sidewalk@hthreeofreqHeter}{26.24}
\newcommand{\building@hthreeofreqHeter}{10.00}
\newcommand{\vegetation@hthreeofreqHeter}{24.02}
\newcommand{\car@hthreeofreqHeter}{2.92}
\newcommand{\truck@hthreeofreqHeter}{0.99}
\newcommand{\bus@hthreeofreqHeter}{0.25}
\newcommand{\twoWheeler@hthreeofreqHeter}{0.07}
\newcommand{\person@hthreeofreqHeter}{0.97}
\newcommand{\pole@hthreeofreqHeter}{0.50}
\newcommand{\hthreeofreqHeter}[1]{{\csname #1@hthreeofreqHeter\endcsname}}
\makeatother

\newcommand{\ColHeadHomo}[3]{\rotatebox{90}{\hspace{-6pt} \textcolor{#1}{$\blacksquare$}\hspace{-6pt}\renewcommand\arraystretch{0.6} \begin{tabular}{l}#2 \\ \hspace{-4pt} ~\tiny(\hthreeofreqHomo{#3}\%) \\ \end{tabular}}}
\newcommand{\ColHeadHeter}[3]{\rotatebox{90}{\hspace{-6pt} \textcolor{#1}{$\blacksquare$}\hspace{-6pt}\renewcommand\arraystretch{0.6} \begin{tabular}{l}#2 \\ \hspace{-4pt} ~\tiny(\hthreeofreqHeter{#3}\%) \\ \end{tabular}}}

\begin{document}
\title{VoxSAOR: Semantic-Aware Occupancy Robustification via Aleatoric Uncertainty Modeling}

\author{Mengfei Duan$^{1}$, Yuheng Zhang$^{1}$, Guoqiang Zhao$^{1}$, Yuanfan Zheng$^{1}$,\\Kailun Yang$^{1,\dag}$, and Zhiyong Li$^{1,\dag}$
\thanks{This work was supported in part by the National Natural Science Foundation of China (Grant No. 62473139 and No. 62388101) and in part by the Hunan Provincial Research and Development Project (Grant No. 2025QK3019).
}
\thanks{$^{1}$The authors are with the School of Artificial Intelligence and Robotics and the National Engineering Research Center of Robot Visual Perception and Control Technology, Hunan University, Changsha, China (e-mail: kailun.yang@hnu.edu.cn; zhiyong.li@hnu.edu.cn).}
\thanks{$^{\dag}$Corresponding authors: Kailun Yang and Zhiyong Li.}
}

\maketitle

\begin{abstract}
Vision-based 3D occupancy prediction provides a dense semantic representation of scenes for autonomous driving and embodied perception, yet remains vulnerable to occlusion, depth ambiguity, and projection errors that induce spatially varying reliability in voxel features. Existing methods typically map voxel representations to semantic predictions without explicitly modeling such observation-induced feature variations. To address this limitation, we propose VoxSAOR, a semantic-aware occupancy robustification framework that models aleatoric uncertainty through feature-space proxies. It first quantifies Local Semantic Consistency (LSC) to derive a voxel-wise structural stability prior and uses cross-view semantic consensus to filter unreliable semantic correspondences. Within each semantic category, spatially matched stable and ambiguous voxels form stable-to-ambiguous feature transitions, which are modeled by a class-conditional diagonal Gaussian distribution, capturing both feature displacement and dispersion. Guided by these proxy distributions, Aleatoric Regime Ensemble Learning (AREL) trains independently parameterized prediction heads using feature perturbations at progressively increasing scales, with perturbation strength modulated by structural reliability. At inference, proxy modeling and sampling are disabled, and all heads process the same observed features; their output probabilities are then averaged to produce a deterministic consensus prediction without requiring sampling. Experiments across multiple datasets and occupancy backbones demonstrate the effectiveness of VoxSAOR in improving semantic occupancy prediction, long-tail recognition, and predictive calibration. The source code will be made publicly available at \url{https://github.com/MengfeiD/VoxSAOR}.
\end{abstract}


\IEEEpeerreviewmaketitle

\section{Introduction}

\begin{figure}
    \centering
    \includegraphics[width=\linewidth]{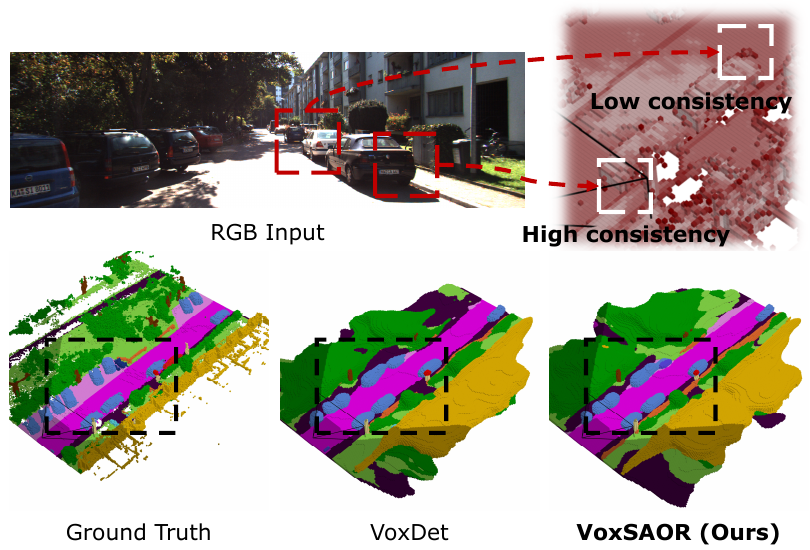}
    \caption{\textbf{Motivation and qualitative illustration} of VoxSAOR. 
    Top: an RGB input and its corresponding Local Semantic Consistency (LSC) visualization, where high- and low-consistency regions indicate structurally stable and locally ambiguous voxels, respectively.
    Red dashed boxes and arrows highlight the corresponding regions.
    Bottom: qualitative comparison of ground truth and predictions from VoxDet~\cite{VoxDet} and VoxSAOR.}
    \label{fig:lscMotivation}
\end{figure}

\IEEEPARstart{V}{ision}-based 3D occupancy prediction aims to recover the geometric occupancy states and semantic composition of a scene from monocular~\cite{MonoScene,VoxFormer,OccFormer,Symphonize} or surround-view images~\cite{surroundOcc,tang2024sparseocc,wang2023openoccupancy}, providing a unified and dense 3D representation of both visible and occluded regions.
It has therefore emerged as an important perception paradigm for autonomous driving~\cite{tong2023scene,SemanticKITTI,tian2023occ3d,li2024sscbench,wang2026occsora}, embodied intelligence~\cite{wu2025embodiedocc,wang2025embodiedocc++,shi2026oneocc,duan2026o3n}, and general 3D scene understanding~\cite{huang2024selfocc,cao2026occany,liu2026occanyscene}.
Despite substantial progress in 2D-to-3D lifting~\cite{MonoScene,philion2020lift,li2024bevformer,chen2025alocc}, spatial feature aggregation~\cite{TPVFormer,SGN,ma2024cotr,huang2024gaussianformer}, and contextual reasoning~\cite{VoxDet,CGFormer,lu2024octreeocc}, inferring a complete 3D scene from limited visual observations remains challenging due to occlusion, depth ambiguity, projection errors, and sensor noise~\cite{VoxFormer,chen2025alocc,roldao20223d,xu2025survey}.

Most existing occupancy prediction methods~\cite{VoxDet,MonoScene,VoxFormer,OccFormer,SGN,CGFormer} learn a deterministic mapping from an observed voxel representation to a semantic output, implicitly assuming that the extracted features provide sufficiently stable evidence for prediction.
This assumption becomes particularly restrictive around semantic and geometric boundaries~\cite{VoxDet,Symphonize}, where reliable features are critical for determining object extent, separating adjacent structures, and distinguishing occupied from free space. 
Recent instance-centric formulations further underscore the importance of such structural cues; for example, VoxDet~\cite{VoxDet} predicts voxel-to-boundary offsets and exploits boundary-associated features for instance-level semantic aggregation. 
However, these informative regions may also be particularly susceptible to observation ambiguity, as occlusion and inaccurate depth lifting can substantially perturb voxel representations near semantic transitions~\cite{VoxFormer,chen2025alocc,CGFormer,chen2025semantic_causality,zhu2026dr}.
Importantly, such ambiguity is distinct from epistemic uncertainty arising from limited knowledge of model parameters.
Instead, it originates from incomplete or ambiguous visual evidence in the available observations and cannot necessarily be eliminated through model optimization alone. These observation-induced variations are associated with aleatoric uncertainty~\cite{kendall2017uncertainties} and may cause semantically or geometrically similar voxels to exhibit different latent representations. 
Ignoring such variability leaves deterministic predictors vulnerable to plausible perturbations of the observed features, thereby reducing their robustness and reliability in complex 3D scenes~\cite{wang2025reliable}. 
Explicitly modeling such aleatoric uncertainty in the latent occupancy space is nevertheless nontrivial. 
Training data typically provide neither paired stable and ambiguous observations of the same voxel nor calibrated voxel-level uncertainty annotations~\cite{SemanticKITTI,tian2023occ3d,li2024sscbench}.
Moreover, feature differences between arbitrary voxels jointly encode semantic identity, object instances, spatial context, and scene layout, making it difficult to isolate observation-induced variations from genuine structural discrepancies.

\begin{figure}[!t]
    \centering
    \includegraphics[width=.9\linewidth]{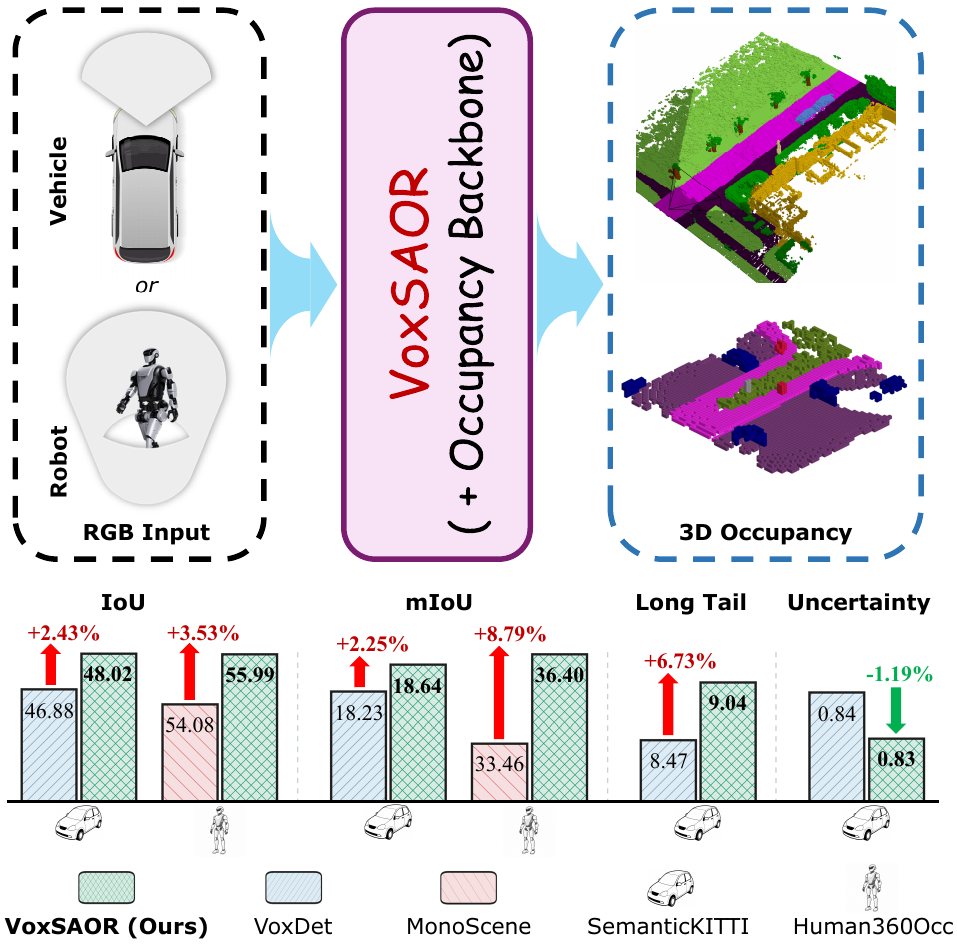}
    \caption{\textbf{Architecture-agnostic applicability and performance improvements} of the proposed VoxSAOR.
    VoxSAOR can be flexibly integrated into different vision-based 3D semantic occupancy prediction backbones~\cite{VoxDet,MonoScene} for autonomous driving and embodied perception, regardless of their voxel aggregation designs.
    Representative comparisons on SemanticKITTI~\cite{SemanticKITTI} and Human360Occ~\cite{shi2026oneocc} public benchmarks show improvements in IoU, mIoU, and long-tail mIoU, along with a reduction in the reported uncertainty-related metric.
    }
    \label{fig:teaserComp}
    \vskip-2ex
\end{figure}

To address this challenge, we propose VoxSAOR, a Semantic-Aware Occupancy Robustification framework via aleatoric uncertainty proxy modeling for robust 3D occupancy prediction.
We first introduce Local Semantic Consistency (LSC) quantification, which converts the local topology of the semantic field into a dense voxel-wise stability prior.
As illustrated in Fig.~\ref{fig:lscMotivation}, high-consistency voxels are treated as relatively stable semantic references, whereas low-consistency voxels indicate local semantic transitions and are more likely to exhibit structural ambiguity.
Cross-view semantic consensus further filters unreliable correspondences using the agreement between occupancy semantic labels and image-space semantic cues projected into the 3D space~\cite{zhang2023simple,samet2026losc}.
Subsequently, structurally stable and ambiguous voxels are matched within each semantic category under spatial constraints, reducing discrepancies caused by semantic identity and scene context.
The resulting stable-to-ambiguous feature transitions are treated as empirical samples of observation-induced latent variation. 
These transitions are modeled by a class-conditional diagonal Gaussian distribution that preserves both dominant direction and stochastic dispersion, yielding an empirically estimable proxy for semantic aleatoric variation.

We further propose Aleatoric Regime Ensemble Learning (AREL) to incorporate the estimated proxy distributions into robust occupancy prediction. 
During training, multiple independently parameterized prediction heads are exposed to progressively different proxy-induced feature regimes, enabling each predictor to adapt to a distinct level of aleatoric feature variation and thereby improving robustness across varying uncertainty regimes. 
The perturbations are modulated by the structural reliability to avoid indiscriminate feature corruption.
At inference, proxy sampling is removed, and all prediction heads process the same observed feature representation. 
Their probability predictions are averaged to form a deterministic predictive consensus.
As shown in Fig.~\ref{fig:teaserComp}, VoxSAOR is independent of specific voxel aggregation operators, allowing it to be flexibly integrated into different vision-based 3D semantic occupancy prediction architectures~\cite{VoxDet,MonoScene,SGN,CGFormer}. 
The consistent improvements in semantic accuracy, long-tail recognition, and predictive reliability across driving and embodied perception scenarios highlight the potential of VoxSAOR to improve the robustness of occupancy predictions.

Our main contributions are summarized as follows:
\begin{itemize}
    \item We propose VoxSAOR, a semantic-aware occupancy robustification framework that models observation-induced aleatoric uncertainty to improve the accuracy and robustness of vision-based 3D occupancy prediction.
    \item We introduce Local Semantic Consistency (LSC) quantification and a stable-to-ambiguous transition construction strategy to derive empirical proxy samples from occupancy supervision. Cross-view semantic consensus, class-wise stratification, and spatial matching jointly mitigate confounding variations arising from unreliable semantic correspondences, semantic identity, and scene context. The resulting transitions are modeled by a class-conditional diagonal Gaussian distribution that characterizes their directional tendency and stochastic dispersion.
    \item We further propose Aleatoric Regime Ensemble Learning (AREL), which exposes independently parameterized prediction heads to progressively different proxy-induced feature regimes through reliability-aware modulation, enabling adaptation to varying degrees of aleatoric feature variation. At inference, all heads process the same observed representation to form a deterministic consensus in probability space.
    \item A comprehensive variety of experiments across multiple occupancy models and datasets, together with ablation studies, demonstrate the effectiveness and generalizability of the proposed framework.
\end{itemize}

\section{Related Work}

\subsection{Vision-based 3D Occupancy Prediction}
Vision-based 3D occupancy prediction aims to infer voxel-wise occupancy and semantic labels from monocular~\cite{VoxDet,MonoScene,VoxFormer,TPVFormer,SGN} or multi-view images~\cite{surroundOcc,chen2025alocc,zhang2026lightocc}.
The field has progressed from dense image-to-volume reconstruction and monocular scene completion~\cite{MonoScene,yao2023ndc,MonoOcc} to multi-view feature aggregation, unified occupancy representations, and efficient global modeling with compact or multi-plane transformers~\cite{VoxFormer,OccFormer,surroundOcc,tong2023scene,TPVFormer,ma2024cotr,wang2024panoocc}.
As reliable 2D-to-3D transformation became a central challenge, later studies increasingly combined depth-aware sparse guidance~\cite{SGN}, stereo geometry~\cite{StereoScene}, contextual instance reasoning~\cite{Symphonize}, joint context-geometry modeling~\cite{CGFormer,xiao2026enhanced}, and visual foundation-model priors~\cite{feng2025vipocc}.
Further advances exploit implicit-explicit depth fusion~\cite{tan2025geocc}, occlusion- or region-aware feature lifting~\cite{chen2025alocc,zhu2026dr}, and object-centric semantic modeling~\cite{wang2026towards} to strengthen 3D feature construction under incomplete visual observations.

In parallel, occupancy representations have shifted from regular dense voxel grids toward adaptive formulations that concentrate computation on informative scene structures.
Sparse voxel methods~\cite{tang2024sparseocc,liu2024fully} reduce redundant dense computation, whereas octree-based methods~\cite{lu2024octreeocc,ouyang2024octocc} organize features hierarchically at varying spatial resolutions.
Point-set and detection-oriented formulations recast exhaustive voxel prediction as compact spatial-element or structured-hypothesis modeling~\cite{VoxDet,shi2024occupancy}, whereas prototype decoding and hierarchical aggregation improve compact structural reasoning~\cite{kim2025protoocc,wang2026hgroupscene}.
Gaussian-based approaches~\cite{huang2024gaussianformer,shi2026odg,park2025s2go,zhou2026generalizing,song2026graphgsocc} further replace discrete grids with continuous learnable primitives for flexible scene representation.
Despite these advances, most existing methods improve feature lifting, representation efficiency, or contextual aggregation within deterministic scene representations~\cite{VoxDet,tang2024sparseocc,SGN,huang2024gaussianformer,CGFormer,ye2024cvt}. 
In contrast, our method characterizes observation-induced aleatoric variations directly in the latent voxel space by deriving an empirical, class-conditional proxy from semantically consistent and spatially matched transitions between structurally stable and ambiguous voxels. This proxy is then leveraged to train prediction heads under progressively scaled, reliability-modulated perturbation regimes, thereby improving the accuracy and robustness of occupancy prediction.

\begin{figure*}[!t]
    \centering
    \includegraphics[width=\textwidth]{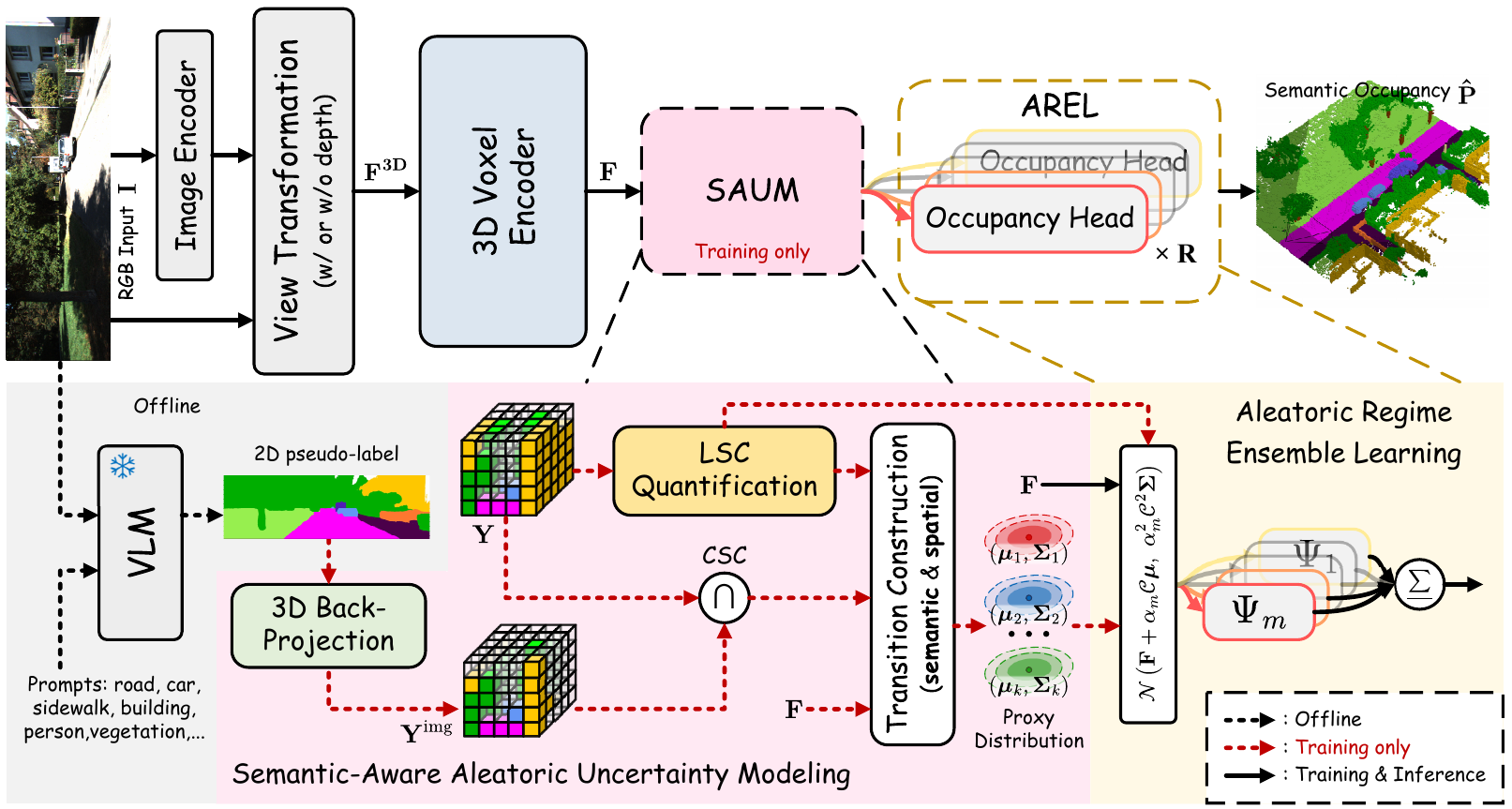}
    \caption{\textbf{Overview of VoxSAOR.} The Semantic-Aware Aleatoric Uncertainty Modeling (SAUM) module constructs class-conditional feature-space proxies using Local Semantic Consistency (LSC) and Cross-View Semantic Consensus (CSC). Guided by these proxies, Aleatoric Regime Ensemble Learning (AREL) trains multiple prediction heads under different perturbation regimes. During inference, proxy modeling and sampling are disabled, and the heads ensemble their predicted class probabilities to produce deterministic, sampling-free semantic occupancy predictions.}
    \label{fig:overview}
\end{figure*}

\subsection{Uncertainty Quantification in Occupancy Prediction}
Uncertainty quantification assesses prediction reliability by distinguishing epistemic uncertainty arising from limited model knowledge from aleatoric uncertainty associated with inherent observation ambiguity~\cite{kendall2017uncertainties}.
In 3D occupancy prediction, existing studies have explored ensemble-based uncertainty estimation through predictive disagreement~\cite{cao2024pasco}, as well as uncertainty characterization through iterative generative predictions~\cite{wang2024occgen}.
Other approaches investigate epistemic uncertainty under camera corruptions and distribution shifts~\cite{heidrich2025occuq}, employ evidential representations to distinguish unobserved regions from free space~\cite{kalble2025evocc}, and model voxel-level predictive confidence using Gaussian distributions alongside geometry-aware feature mapping~\cite{liu2025geometric_mapping}.
Further developments incorporate hybrid voxel-level uncertainty modeling and confidence calibration~\cite{wang2025reliable}, propagate depth uncertainty into occupancy prediction with coverage-guaranteed conformal prediction~\cite{su2026alpha}, and exploit uncertainty-guided sampling for online scene updates~\cite{wang2025embodiedocc++}.
Cross-modal alignment has also been explored to reduce semantic uncertainty in occupancy prediction~\cite{bea2026high_fidelity}.
Collectively, these studies extend uncertainty-aware occupancy prediction toward diverse aspects of uncertainty estimation, representation, calibration, propagation, and reliability-oriented learning.

Nevertheless, existing methods primarily quantify uncertainty through output distributions, repeated predictions, or evidence parameters, without explicitly modeling observation-induced variations in voxel features.
To address this gap, our method derives a class-conditional aleatoric proxy from structurally constrained stable-to-ambiguous feature transitions and leverages it to guide reliability-aware multi-regime learning. This feature-space formulation improves semantic occupancy robustness without repeated stochastic sampling.

\section{Methodology}

\subsection{Overview}
As illustrated in Fig.~\ref{fig:overview}, our proposed VoxSAOR augments a generic vision-based 3D occupancy predictor~\cite{VoxDet,MonoScene,shi2026oneocc,SGN,CGFormer} with two complementary components: Semantic-Aware Aleatoric Uncertainty Modeling (SAUM, Sec.~\ref{sec:sau}) and Aleatoric Regime Ensemble Learning (AREL, Sec.~\ref{sec:arel}). 
Given an RGB Input $\mathbf{I}$, an image encoder, a view transformation module, and a 3D voxel encoder first extract the voxel feature field $\mathbf{F}$.
During training, SAUM constructs feature-space proxies to characterize observation-induced aleatoric variations using standard occupancy supervision. 
Specifically, Local Semantic Consistency Quantification (LSC, Sec.~\ref{sec:lsc}) derives a voxel-wise stability prior from the local topology of the semantic occupancy field, while cross-view semantic consensus filters unreliable candidates using projected image-space semantic cues~\cite{zhang2023simple,samet2026losc}.
Within each semantic category, structurally stable voxels are matched with ambiguous voxels subject to spatial constraints, yielding stable-to-ambiguous feature transitions. These transitions are subsequently modeled using class-conditional diagonal Gaussian distributions, which characterize the corresponding feature displacements and degrees of dispersion.

AREL further incorporates these proxy distributions into occupancy learning. 
During training, transition samples are drawn from the distributions and used to perturb the observed voxel features at multiple scales, thereby exposing independently parameterized prediction heads to distinct aleatoric regimes. 
All prediction heads are trained under the same occupancy supervision and learn to preserve semantic predictions under plausible feature variations. 
At inference, SAUM and proxy sampling are disabled, incurring no additional inference overhead from uncertainty modeling. All heads operate directly on the same observed feature field, and their predictions are averaged to obtain the final semantic occupancy output $\mathbf{\hat{P}}$.

\subsection{Local Semantic Consistency Quantification}
\label{sec:lsc}

\begin{figure}
    \centering
    \includegraphics[width=\linewidth]{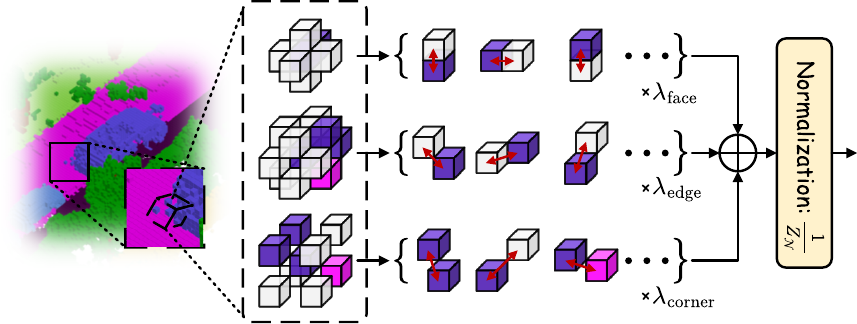}
    \caption{\textbf{Local Semantic Consistency (LSC) quantification.} Semantic agreement with face-, edge-, and corner-connected neighbors is weighted and normalized to derive a voxel-wise structural stability prior.}
    \label{fig:lsc}
\end{figure}

Local semantic consistency provides a structural cue for characterizing voxel reliability. 
Voxels inside homogeneous regions are generally supported by consistent contextual evidence, whereas those near object boundaries, thin structures, or fragmented regions tend to exhibit more ambiguous representations. 
Motivated by this observation, we quantify \textit{Local Semantic Consistency} (LSC), which converts the discrete topology of a semantic field into a dense voxel-wise stability prior, as illustrated in
Fig.~\ref{fig:lsc}.

Let $\mathbf{Y}\in\mathbb{L}^{H\times W\times D}$ denote a semantic volume aligned with the voxel features. Here, $\mathbb{L}$ is the semantic label space, and $y_p$ denotes the semantic label at voxel position $p$. 
Given a local 3D neighborhood $\mathcal{N}$, such as $6$-, $18$-, or $26$-connectivity, we define the LSC score as the normalized weighted agreement between the center voxel and its valid neighbors:
\begin{equation}
\mathcal{C}(p) = \frac{1}{Z_{\mathcal{N}}}
\sum_{\delta\in\mathcal{N}}
w(\delta)\,
\mathbb{I}
\left[
y_p=y_{p+\delta}
\right]
M(p,\delta),
\label{eq:lsc}
\end{equation}
where $\mathcal{C}(p)\in[0,1]$, $\mathbb{I}[\cdot]$ denotes the indicator function, $M(p,\delta)$ is a binary validity mask indicating whether $p+\delta$ lies inside the semantic volume, and $Z_{\mathcal{N}}$ is the maximum weighted agreement under the selected neighborhood.

To distinguish different spatial relationships, we employ a connectivity-aware weighting function:
\begin{equation}
w(\delta)=
\begin{cases}
\lambda_{\mathrm{face}},
& \|\delta\|_1=1,\\
\lambda_{\mathrm{edge}},
& \|\delta\|_1=2,\\
\lambda_{\mathrm{corner}},
& \|\delta\|_1=3.
\end{cases}
\label{eq:lsc_weight}
\end{equation}
Here, $\lambda_{\mathrm{face}}$, $\lambda_{\mathrm{edge}}$, and $\lambda_{\mathrm{corner}}$ independently control the contributions of face-, edge-, and corner-connected neighbors. 
Accordingly, the normalization factor is defined as:
\begin{equation}
Z_{\mathcal N}=\sum_{\delta\in\mathcal N}w(\delta).
\label{eq:lsc_normalization}
\end{equation}
A high LSC score indicates strong local semantic agreement and thus greater structural stability, whereas a low score indicates local semantic discontinuities and a higher degree of structural ambiguity.

For efficient computation, we construct a bank of $3\times3\times3$ sparse kernels, each selecting one neighboring offset in $\mathcal N$. A single 3D convolution extracts all neighboring label volumes in parallel. The center labels are then broadcast and compared with the neighboring labels, while directional boundary masks exclude out-of-range positions. The resulting semantic agreements are weighted and reduced along the neighborhood dimension to obtain the dense LSC map: 
\begin{equation}
\mathbf{C}_{local}
=
\frac{1}{Z_{\mathcal{N}}}
\operatorname{ReduceSum}_{\mathcal{N}}
\left(
\mathbb{I}
\left[
\mathbf{Y}
=
\mathbf{Y}_{\mathcal{N}}
\right]
\odot
\mathbf{W}_{\mathcal{N}}
\odot
\mathbf{M}_{\mathcal{N}}
\right).
\label{eq:lsc_parallel}
\end{equation}
Here, $\odot$ denotes element-wise multiplication, $\mathbf{W}_{\mathcal{N}}$ contains connectivity-aware weights, and $\mathbf{M}_{\mathcal{N}}$ contains the directional boundary-validity masks.
Unlike predictive entropy or confidence-based uncertainty, LSC is derived from the local topology of the semantic field rather than model predictions.
It thus provides a structure-driven stability prior that does not directly depend on predictive confidence or calibration.

\subsection{Semantic-Aware Aleatoric Uncertainty Modeling}
\label{sec:sau}

Vision-based 3D occupancy prediction requires recovering a complete 3D semantic field from limited 2D observations, resulting in voxel features with spatially varying reliability. Occlusion, depth ambiguity, and limited visibility during 2D-to-3D lifting~\cite{MonoScene,VoxFormer,OccFormer} can alter the visual evidence available for voxel representation learning, leading to observation-dependent feature variations. Such variations are related to aleatoric uncertainty, which arises from inherent ambiguity in the available observations~\cite{kendall2017uncertainties}.
However, feature discrepancies between voxels cannot be attributed exclusively to observation-induced aleatoric uncertainty, as they may also reflect differences in semantic context, object instances, spatial structure, or imperfect model representations. Directly modeling arbitrary feature differences may therefore introduce substantial confounding variations. Rather than attempting to recover a calibrated predictive uncertainty distribution or isolate the exact aleatoric component, we construct a structured empirical proxy using stable-to-ambiguous feature transitions.

For a voxel $p$ at spatial coordinate $\mathbf{x}_p$ belonging to semantic category $k$, we conceptually decompose its feature as
\begin{equation}
\mathbf{f}_p
=
\overline{\mathbf{f}}_k(\mathbf{x}_p)
+
\boldsymbol{\eta}_p
+
\mathbf{r}_p,
\label{eq:feature_decomposition}
\end{equation}
where $\overline{\mathbf{f}}_k(\mathbf{x}_p)$ denotes the locally expected semantic representation conditioned on category $k$ and spatial context $\mathbf{x}_p$, $\boldsymbol{\eta}_p$ denotes the feature variation induced by observation ambiguity, and $\mathbf{r}_p$ collects nuisance variations that cannot be explicitly separated, such as model-dependent representation errors. Our objective is to construct transition samples enriched in $\boldsymbol{\eta}_p$ while suppressing the other components through semantic and spatial constraints.

\subsubsection{Cross-View Semantic Consensus}
Estimating a transition distribution from all labeled voxels may introduce unreliable samples caused by inaccurate image-to-voxel correspondence or inconsistent semantic evidence. We therefore construct a reliable candidate set using the agreement between voxel-wise semantic labels and image-space semantic cues projected into the 3D voxel space. During training, the image-space semantic cues are obtained from offline 2D pseudo-labels generated by a pretrained open-vocabulary segmentation model~\cite{radford2021learning,zhang2023simple}. Let $\hat{y}^{\,\mathrm{img}}_p$ denote the resulting image-space semantic label projected onto voxel $p$. For each non-empty semantic category $k$, the semantic consensus set is defined as: 
\begin{equation}
\Omega_k
=
\left\{
p
\;\middle|\;
y_p=k,\;
\hat{y}^{\,\mathrm{img}}_p=k
\right\}.
\label{eq:semantic_consensus}
\end{equation}

\subsubsection{Stable-to-Ambiguous Transition Construction}
The LSC score in Sec.~\ref{sec:lsc} is used as a structural ranking criterion to distinguish relatively stable and ambiguous voxel candidates.
High-consistency voxels are treated as relatively stable references, whereas low-consistency voxels are treated as candidates more likely to exhibit observation-induced semantic ambiguity.
To avoid unreliable anchor construction for sparsely represented categories, we retain only categories with sufficient consensus support:
\begin{equation}
\mathcal{K}_{\mathrm{valid}}
=
\left\{
k
\;\middle|\;
|\Omega_k|
\geq
n_{\min}
\right\},
\label{eq:valid_semantic_categories}
\end{equation}
where $n_{\min}$ denotes the minimum number of candidate voxels required for reliable anchor construction. Categories outside $\mathcal{K}_{\mathrm{valid}}$ are skipped in the current batch and introduce no proxy-induced perturbation.

For each semantic category $k\in\mathcal{K}_{\mathrm{valid}}$, we randomly select the $n_a$ voxels with the lowest LSC scores as ambiguous anchors:
\begin{equation}
\mathcal{A}_k
=
\underset{p\in\Omega_k}
{\operatorname{BottomK}_{n_a}}
\;
\mathcal{C}(p).
\label{eq:ambiguous_anchor}
\end{equation}
A larger stable candidate pool is constructed from high-consistency voxels while excluding the ambiguous anchors:
\begin{equation}
\mathcal{S}_k
=
\underset{p\in\Omega_k\setminus\mathcal{A}_k}
{\operatorname{TopK}_{n_s}}
\;
\mathcal{C}(p),
~n_s\geq n_a.
\label{eq:stable_anchor}
\end{equation}
Using multiple anchors captures diverse stable-to-ambiguous transitions and reduces sensitivity to extreme samples.

To mitigate contextual and instance-level differences, each ambiguous anchor $a\in\mathcal{A}_k$ is matched to its spatially nearest stable candidate within the same semantic category:
\begin{equation}
\pi_k(a)
=
\arg\min_{s\in\mathcal{S}_k}
\left\|
\mathbf{x}_a-\mathbf{x}_s
\right\|_2.
\label{eq:spatial_anchor_matching}
\end{equation}

For each matched pair, we define the stable-to-ambiguous transition vector as
\begin{equation}
\boldsymbol{\Delta}_{k,a}
=
\mathbf{f}_a
-
\mathbf{f}_{\pi_k(a)}.
\label{eq:transition_vector}
\end{equation}
Substituting Eq.~\eqref{eq:feature_decomposition} into Eq.~\eqref{eq:transition_vector} yields
\begin{align}
\boldsymbol{\Delta}_{k,a}
&=
\underbrace{
\overline{\mathbf{f}}_k(\mathbf{x}_a)
-
\overline{\mathbf{f}}_k
\left(
\mathbf{x}_{\pi_k(a)}
\right)
}_{\Delta\overline{\mathbf{f}}_{k,a}}
+
\underbrace{
\boldsymbol{\eta}_a
-
\boldsymbol{\eta}_{\pi_k(a)}
}_{\Delta\boldsymbol{\eta}_{k,a}}
+
\underbrace{
\mathbf{r}_a
-
\mathbf{r}_{\pi_k(a)}
}_{\Delta\mathbf{r}_{k,a}}.
\label{eq:transition_decomposition}
\end{align}
Although the transition contains residual semantic-context and representation differences, same-category spatial matching mitigates these confounding factors, while LSC-based stratification emphasizes differences associated with structural ambiguity. We therefore use these transitions as proxies for observation-induced aleatoric variations.

\subsubsection{Class-Conditional Aleatoric Proxy Distribution}
For category $k$, we collect all valid transition vectors into
\begin{equation}
\mathcal{D}_k
=
\left\{
\boldsymbol{\Delta}_{k,a}
\;\middle|\;
a\in\mathcal{A}_k
\right\}.
\label{eq:transition_set}
\end{equation}
Since semantic categories may exhibit distinct ambiguity patterns due to differences in geometry, scale, appearance, and visibility, we estimate a separate transition distribution for each category. The channel-wise empirical mean and variance are given by
\begin{equation}
\begin{aligned}
\boldsymbol{\mu}_k
&=
\frac{1}{|\mathcal{D}_k|}
\sum_{\boldsymbol{\Delta}\in\mathcal{D}_k}
\boldsymbol{\Delta},\\
\boldsymbol{\sigma}_k^2
&=
\frac{1}{|\mathcal{D}_k|}
\sum_{\boldsymbol{\Delta}\in\mathcal{D}_k}
(\boldsymbol{\Delta}-\boldsymbol{\mu}_k)
\odot
(\boldsymbol{\Delta}-\boldsymbol{\mu}_k).
\end{aligned}
\end{equation}
The mean $\boldsymbol{\mu}_k$ describes the dominant direction of the selected stable-to-ambiguous transitions, while $\boldsymbol{\sigma}_k$ characterizes their stochastic dispersion across feature channels.

For computational tractability, we approximate the class-conditional transition distribution with a diagonal Gaussian, yielding the semantic-aware aleatoric proxy distribution:
\begin{equation}
\begin{aligned}
\boldsymbol{\Sigma}_k
&:=
\operatorname{diag}
\left(
\boldsymbol{\sigma}_k^2
\right),\\
\widetilde{q}
\left(
\boldsymbol{\Delta}\mid k
\right)
&:=
\mathcal{N}
\left(
\boldsymbol{\Delta};
\boldsymbol{\mu}_k,\,
\boldsymbol{\Sigma}_k
\right).
\end{aligned}
\label{eq:aleatoric_proxy_distribution}
\end{equation}

\subsection{Aleatoric Regime Ensemble Learning}
\label{sec:arel}
Voxel features reconstructed from visual observations may contain observation-induced variations associated with occlusion, projection errors, and semantic ambiguity. Rather than recovering an inaccessible clean representation, we aim to improve prediction robustness under plausible feature variations around the observed voxel representation. 
Previous studies have demonstrated the effectiveness of multi-input multi-output (MIMO) ensembles for robust prediction and uncertainty estimation~\cite{cao2024pasco,havasi2021training}.
Unlike conventional multi-input ensembles, we propose Aleatoric Regime Ensemble Learning (AREL), which trains multiple prediction heads under distinct perturbation regimes derived from class-conditional aleatoric uncertainty proxies.

\subsubsection{Reliability-Constrained Proxy Realization}
Given the voxel feature field $\mathbf{F}$ with voxel-wise representation $\mathbf{f}_p$, AREL constructs $R$ perturbation branches with progressively increasing scales:
\begin{equation}
0 < \alpha_1 < \alpha_2 < \cdots < \alpha_R,
\label{eq:arel_perturbed_scales}
\end{equation}
where $\alpha_m=m\alpha$ and $\alpha>0$ is the base perturbation strength.

During training, class-conditional proxy sampling is guided by the available semantic occupancy labels. For voxels belonging to categories with valid proxy estimates, denoted by $\mathcal{V}_{\mathrm{proxy}}=\{p\mid y_p\in\mathcal{K}_{\mathrm{valid}}\}$, each branch independently samples a transition vector from the distribution:
\begin{equation}
\boldsymbol{\Delta}^{(m)}_p
\overset{\mathrm{ind}}{\sim}
\widetilde{q}
\left(
\cdot\mid y_p
\right),
\quad
p\in\mathcal{V}_{\mathrm{proxy}},
\quad m=1,\ldots,R.
\label{eq:arel_perturbed_sampling}
\end{equation}
The perturbed voxel representation is defined as
\begin{equation}
\mathbf{f}^{(m)}_p
=
\begin{cases}
\mathbf{f}_p+
\alpha_m\mathcal{C}(p)
\boldsymbol{\Delta}^{(m)}_p,
& p\in\mathcal{V}_{\mathrm{proxy}},\\
\mathbf{f}_p,
& \text{otherwise},
\end{cases}
\label{eq:arel_perturbed_features}
\end{equation}
where $\mathcal{C}(p)$ denotes the normalized LSC score. The resulting branch feature field is denoted by $\mathbf{F}^{(m)}=\{\mathbf{f}^{(m)}_p\}_p$.
Specifically, perturbations are attenuated in low-consistency regions and emphasized in structurally stable regions. Although ambiguous voxels contribute to proxy estimation, applying strong perturbations directly to their potentially unreliable representations may introduce additional semantic distortion. 
Using relatively stable features as perturbation carriers instead encourages the predictor to preserve semantic decisions under controlled observation-related variations.

\subsubsection{Multi-Regime Uncertainty Learning}
Each perturbed feature field is processed by a separately parameterized semantic occupancy head:
\begin{equation}
\begin{aligned}
\mathbf{Z}_m
&=
\Psi_m
\left(
\mathbf{F}^{(m)};
\boldsymbol{\theta}_m
\right),\\
\mathbf{P}_m
&=
\operatorname{Softmax}
\left(
\mathbf{Z}_m
\right),
\end{aligned}
\label{eq:arel_perturbed_prediction}
\end{equation}
where all branches share the same feature extraction backbone but employ separate prediction parameters $\{\boldsymbol{\theta}_m\}_{m=1}^{R}$.
By assigning distinct perturbation scales to different prediction heads, AREL exposes them to progressively stronger proxy-induced feature variations. 
All branches are optimized using the same semantic occupancy supervision:
\begin{equation}
\mathcal{L}
=
\frac{1}{R}
\sum_{m=1}^{R}
\mathcal{L}_{\mathrm{occ}}
\left(
\mathbf{P}_m,\mathbf{Y}
\right),
\label{eq:arel_perturbed_supervision}
\end{equation}
where $\mathcal{L}_{\mathrm{occ}}$ denotes the occupancy prediction objective. The common supervision constrains every branch to preserve the underlying semantic prediction across its corresponding uncertainty neighborhood.

\subsubsection{Observed Feature Prediction}
At inference, all training-only SAUM operations, including external semantic cue processing and proxy construction, as well as perturbations, are disabled. All prediction heads directly process the same unperturbed voxel feature field $\mathbf{F}$, and the final semantic occupancy prediction is obtained by averaging their predicted class probabilities:
\begin{equation}
\begin{aligned}
\mathbf{P}^{\mathrm{test}}_m
&=
\operatorname{Softmax}
\left(
\Psi_m
\left(
\mathbf{F};
\boldsymbol{\theta}_m
\right)
\right),\\
\widehat{\mathbf{P}}
&=
\frac{1}{R}
\sum_{m=1}^{R}
\mathbf{P}^{\mathrm{test}}_m.
\end{aligned}
\label{eq:arel_perturbed_consensus}
\end{equation}
In this way, the aleatoric proxy guides robust prediction learning during training, while its effect is retained in the learned prediction heads without introducing additional modeling or stochastic sampling at inference.

\begin{table*}[!t]
\newcommand{\clsname}[2]{
\rotatebox{90}{
\hspace{-6pt}
\textcolor{#2}{$\blacksquare$}
\hspace{-6pt}
\renewcommand\arraystretch{0.6}
\begin{tabular}{l}
#1                                      \\
\hspace{-4pt} ~\tiny(\semkitfreq{#2}\%) \\
\end{tabular}
}}
\centering
\caption{\textbf{Quantitative results on SemanticKITTI hidden test set}~\cite{SemanticKITTI}\textbf{.}
\textbf{T} indicates using extra temporal information.
The best and the second-best results are in \textbf{bold} and \underline{underlined}, respectively. R-50 indicates ResNet-50, and Eff-B7 is the stronger EfficientNet-B7 backbone with more parameters. $\ast$ denotes results reproduced using the official code implementation.
}
\label{tab:sem_kitti_test}
\resizebox{\textwidth}{!}{
\begin{tabular}{l|c|c|cc|*{19}{c}}
    \toprule
    Method & \textbf{Arch.} & \textbf{T} & IoU & mIoU & \clsname{road}{road} & \clsname{sidewalk}{sidewalk} & \clsname{parking}{parking} & \clsname{other-grnd.}{otherground} & \clsname{building}{building} & \clsname{car}{car} & \clsname{truck}{truck} & \clsname{bicycle}{bicycle} & \clsname{motorcycle}{motorcycle} & \clsname{other-veh.}{othervehicle} & \clsname{vegetation}{vegetation} & \clsname{trunk}{trunk} & \clsname{terrain}{terrain} & \clsname{person}{person} & \clsname{bicyclist}{bicyclist} & \clsname{motorcyclist}{motorcyclist} & \clsname{fence}{fence} & \clsname{pole}{pole} & \clsname{traf.-sign}{trafficsign} \\
    \midrule
    MonoScene~\cite{MonoScene} & Eff-B7 & & 34.16 & 11.08 & 54.70 & 27.10 & 24.80 & 5.70 & 14.40 & 18.80 & 3.30 & 0.50 & 0.70 & 4.40 & 14.90 & 2.40 & 19.50 & 1.00 & 1.40 & 0.40 & 11.10 & 3.30 & 2.10 \\
    TPVFormer~\cite{TPVFormer} & Eff-B7 & & 34.25 & 11.26 & 55.10 & 27.20 & 27.40 & 6.50 & 14.80 & 19.20 & 3.70 & 1.00 & 0.50 & 2.30 & 13.90 & 2.60 & 20.40 & 1.10 & 2.40 & 0.30 & 11.00 & 2.90 & 1.50 \\
    SurroundOcc~\cite{surroundOcc} & Eff-B7 & & 34.72 & 11.86 & 56.90 & 28.30 & 30.20 & 6.80 & 15.20 & 20.60 & 1.40 & 1.60 & 1.20 & 4.40 & 14.90 & 3.40 & 19.30 & 1.40 & 2.00 & 0.10 & 11.30 & 3.90 & 2.40 \\
    OccFormer~\cite{OccFormer} & Eff-B7 & & 34.53 & 12.32 & 55.90 & 30.30 & {31.50} & 6.50 & 15.70 & 21.60 & 1.20 & 1.50 & 1.70 & 3.20 & 16.80 & 3.90 & 21.30 & 2.20 & 1.10 & 0.20 & 11.90 & 3.80 & 3.70 \\
    IAMSSC~\cite{IAMSSC} & R-50 & & 43.74 & 12.37 & 54.00 & 25.50 & 24.70 & 6.90 & 19.20 & 21.30 & 3.80 & 1.10 & 0.60 & 3.90 & 22.70 & 5.80 & 19.40 & 1.50 & 2.90 & 0.50 & 11.90 & 5.30 & 4.10 \\
    VoxFormer~\cite{VoxFormer} & R-50 & & 42.95 & 12.20 & 53.90 & 25.30 & 21.10 & 5.60 & 19.80 & 20.80 & 3.50 & 1.00 & 0.70 & 3.70 & 22.40 & 7.50 & 21.30 & 1.40 & 2.60 & 0.20 & 11.10 & 5.10 & 4.90 \\
    VoxFormer~\cite{VoxFormer} & R-50 & \checkmark & 43.21 & 13.41 & 54.10 & 26.90 & 25.10 & 7.30 & 23.50 & 21.70 & 3.60 & 1.90 & 1.60 & 4.10 & 24.40 & 8.10 & 24.20 & 1.60 & 1.10 & 0.00 & 13.10 & 6.60 & 5.70 \\
    DepthSSC~\cite{DepthSSC} & R-50 & & {44.58} & 13.11 & 55.64 & 27.25 & 25.72 & 5.78 & 20.46 & 21.94 & 3.74 & 1.35 & 0.98 & 4.17 & 23.37 & 7.64 & 21.56 & 1.34 & 2.79 & 0.28 & 12.94 & 5.87 & 6.23 \\
    Symphonies~\cite{Symphonize} & R-50 & & 42.19 & 15.04 & 58.40 & 29.30 & 26.90 & {11.70} & {24.70} & 23.60 & 3.20 & 3.60 & {2.60} & \SecondBest{5.60} & 24.20 & 10.00 & 23.10 & \SecondBest{3.20} & 1.90 & \Best{2.00} & 16.10 & {7.70} & 8.00 \\
    HASSC~\cite{HASSC} & R-50 & & 43.40 & 13.34 & 54.60 & 27.70 & 23.80 & 6.20 & 21.10 & 22.80 & 4.70 & 1.60 & 1.00 & 3.90 & 23.80 & 8.50 & 23.30 & 1.60 & {4.00} & 0.30 & 13.10 & 5.80 & 5.50 \\
    HASSC~\cite{HASSC} & R-50 & \checkmark & 42.87 & 14.38 & 55.30 & 29.60 & 25.90 & 11.30 & 23.10 & 23.00 & 2.90 & 1.90 & 1.50 & 4.90 & 24.80 & 9.80 & 26.50 & 1.40 & 3.00 & 0.00 & 14.30 & 7.00 & 7.10 \\
    StereoScene~\cite{StereoScene} & Eff-B7 & & 43.34 & 15.36 & {61.90} & {31.20} & 30.70 & 10.70 & 24.20 & 22.80 & 2.80 & 3.40 & {2.40} & \Best{6.10} & 23.80 & 8.40 & {27.00} & {2.90} & 2.20 & 0.50 & 16.50 & 7.00 & 7.20 \\
    H2GFormer~\cite{H2GFormer} & R-50 & & 44.20 & 13.72 & 56.40 & 28.60 & 26.50 & 4.90 & 22.80 & 23.40 & 4.80 & 0.80 & 0.90 & 4.10 & 24.60 & 9.10 & 23.80 & 1.20 & 2.50 & 0.10 & 13.30 & 6.40 & 6.30 \\
    H2GFormer~\cite{H2GFormer} & R-50 & \checkmark & 43.52 & 14.60 & 57.90 & 30.40 & 30.00 & 6.90 & 24.00 & 23.70 & {5.20} & 0.60 & 1.20 & 5.00 & {25.20} & {10.70} & 25.80 & 1.10 & 0.10 & 0.00 & 14.60 & 7.50 & {9.30} \\
    MonoOcc~\cite{MonoOcc} & R-50 & & - & 13.80 & 55.20 & 27.80 & 25.10 & 9.70 & 21.40 & 23.20 & {5.20} & 2.20 & 1.50 & 5.40 & 24.00 & 8.70 & 23.00 & 1.70 & 2.00 & 0.20 & 13.40 & 5.80 & 6.40 \\
    L2COcc-C~\cite{wang2025l2cocc} & Eff-B7 & & 44.31 & 17.03 & \Best{66.00} & 35.00 & 33.10 & \Best{13.50} & 25.10 & \SecondBest{27.20} & 3.00 & 3.50 & 3.60 & 4.30 & 25.20 & 11.50 & 30.10 & 1.50 & 2.40 & 0.20 & 20.50 & {9.10} & 8.90 \\
    HTCL~\cite{li2024hierarchical} & Eff-B7 & \checkmark & 44.23 & {17.09} & {64.40} & 34.80 & 33.80 & {12.40} & {25.90} & \Best{27.30} & \SecondBest{5.70} & 1.80 & 2.20 & 5.40 & {25.30} & 10.80 & {31.20} & 1.10 & 3.10 & 0.90 & \SecondBest{21.10} & {9.00} & 8.30 \\
    
    \midrule
    SGN~\cite{SGN} & R-50 & & 41.88 & 14.01 & 57.80 & 29.20 & 27.70 & 5.20 & 23.90 & 24.90 & 2.70 & 0.40 & 0.30 & 4.00 & 24.20 & 10.00 & 25.80 & 1.10 & 2.50 & 0.30 & 14.20 & 7.40 & 4.40 \\
    {\cellcolor{mygreen!10}}VoxSAOR (+SGN) & {\cellcolor{mygreen!10}}R-50 & {\cellcolor{mygreen!10}} 
    & {\cellcolor{mygreen!10}}42.85 
    & {\cellcolor{mygreen!10}}14.37 
    & {\cellcolor{mygreen!10}}57.40 
    & {\cellcolor{mygreen!10}}29.00 
    & {\cellcolor{mygreen!10}}27.00 
    & {\cellcolor{mygreen!10}}3.90 
    & {\cellcolor{mygreen!10}}25.00 
    & {\cellcolor{mygreen!10}}24.40 
    & {\cellcolor{mygreen!10}}4.60 
    & {\cellcolor{mygreen!10}}1.20 
    & {\cellcolor{mygreen!10}}1.00 
    & {\cellcolor{mygreen!10}}1.80 
    & {\cellcolor{mygreen!10}}24.40 
    & {\cellcolor{mygreen!10}}10.60 
    & {\cellcolor{mygreen!10}}26.40 
    & {\cellcolor{mygreen!10}}2.00 
    & {\cellcolor{mygreen!10}}2.10 
    & {\cellcolor{mygreen!10}}0.10 
    & {\cellcolor{mygreen!10}}16.00 
    & {\cellcolor{mygreen!10}}9.10 
    & {\cellcolor{mygreen!10}}7.10 \\
    
    \midrule
    CGFormer~\cite{CGFormer} & Eff-B7 & & 44.41 & {16.63} & {64.30} & {34.20} & {34.10} & {12.10} & {25.80} & {26.10} & 4.30 & {3.70} & 1.30 & 2.70 & 24.50 & {11.20} & {29.30} & 1.70 & {3.60} & 0.40 & {18.70} & {8.70} & {9.30} \\
    {\cellcolor{mygreen!10}}VoxSAOR (+CGFormer) & {\cellcolor{mygreen!10}}Eff-B7 & {\cellcolor{mygreen!10}} 
    & {\cellcolor{mygreen!10}}44.34 
    & {\cellcolor{mygreen!10}}17.05 
    & {\cellcolor{mygreen!10}}\SecondBest{65.40} 
    & {\cellcolor{mygreen!10}}34.90 
    & {\cellcolor{mygreen!10}}33.80 
    & {\cellcolor{mygreen!10}}11.90 
    & {\cellcolor{mygreen!10}}25.70 
    & {\cellcolor{mygreen!10}}26.10 
    & {\cellcolor{mygreen!10}}5.00 
    & {\cellcolor{mygreen!10}}3.80 
    & {\cellcolor{mygreen!10}}3.80 
    & {\cellcolor{mygreen!10}}5.00 
    & {\cellcolor{mygreen!10}}24.70 
    & {\cellcolor{mygreen!10}}11.20 
    & {\cellcolor{mygreen!10}}28.40 
    & {\cellcolor{mygreen!10}}1.60 
    & {\cellcolor{mygreen!10}}\SecondBest{4.60} 
    & {\cellcolor{mygreen!10}}0.00 
    & {\cellcolor{mygreen!10}}20.30 
    & {\cellcolor{mygreen!10}}8.50 
    & {\cellcolor{mygreen!10}}9.10 \\
    
    \midrule
    VoxDet$^\ast$~\cite{VoxDet} & R-50 & & \SecondBest{46.88} & \SecondBest{18.23} & 64.90 & \SecondBest{35.40} & \SecondBest{34.70} & 13.30 & \SecondBest{28.50} & 27.00 & 4.60 & \Best{6.30} & \Best{5.30} & 4.90 & \SecondBest{28.00} & \SecondBest{13.30} & \SecondBest{31.70} & 2.70 & 3.30 & 0.70 & \Best{21.50} & \SecondBest{10.10} & \SecondBest{10.10} \\
    {\cellcolor{mygreen!10}}VoxSAOR (+VoxDet) & {\cellcolor{mygreen!10}}R-50 & {\cellcolor{mygreen!10}} 
    & {\cellcolor{mygreen!10}}\Best{48.02}
    & {\cellcolor{mygreen!10}}\Best{18.64}
    & {\cellcolor{mygreen!10}}65.10
    & {\cellcolor{mygreen!10}}\Best{35.70}
    & {\cellcolor{mygreen!10}}\Best{35.50}
    & {\cellcolor{mygreen!10}}\SecondBest{13.40}
    & {\cellcolor{mygreen!10}}\Best{28.60}
    & {\cellcolor{mygreen!10}}26.80
    & {\cellcolor{mygreen!10}}\Best{6.30}
    & {\cellcolor{mygreen!10}}\SecondBest{4.90}
    & {\cellcolor{mygreen!10}}\SecondBest{4.80}
    & {\cellcolor{mygreen!10}}4.50
    & {\cellcolor{mygreen!10}}\Best{29.80}
    & {\cellcolor{mygreen!10}}\Best{13.40}
    & {\cellcolor{mygreen!10}}\Best{32.00}
    & {\cellcolor{mygreen!10}}\Best{3.90}
    & {\cellcolor{mygreen!10}}\Best{5.70}
    & {\cellcolor{mygreen!10}}\SecondBest{1.00}
    & {\cellcolor{mygreen!10}}21.00
    & {\cellcolor{mygreen!10}}\Best{10.40}
    & {\cellcolor{mygreen!10}}\Best{11.30} \\
    \bottomrule
\end{tabular}}
\end{table*}

\section{Experiments}

We evaluate VoxSAOR across multiple datasets and occupancy backbones, covering both autonomous driving and embodied perception scenarios.

\subsection{Datasets and Metrics}
\subsubsection{Datasets}
For large-scale outdoor driving scene understanding, we evaluate our method on SemanticKITTI~\cite{SemanticKITTI} and SSCBench-KITTI-360~\cite{li2024sscbench}.
Built upon the KITTI odometry dataset~\cite{geiger2012we}, which comprises $22$ driving sequences, SemanticKITTI~\cite{SemanticKITTI} provides dense semantic annotations and serves as a widely adopted benchmark for 3D occupancy prediction.
Following the official dataset split, sequences $00{-}07$ and $09{-}10$ are used for training with $3,834$ scans, sequence $08$ is used for validation with $815$ scans, and sequences $11{-}21$ are used for testing with $3,901$ scans.
SSCBench-KITTI-360~\cite{li2024sscbench} is a large-scale semantic occupancy benchmark constructed from KITTI-360 with $9$ urban driving sequences. 
Specifically, sequences $00$, $02{-}05$, $07$, and $10$ constitute the training set with $8,487$ frames, while sequence $06$ is used for validation with $1,812$ frames and sequence $09$ for testing with $2,566$ frames. 
For both benchmarks, the volume of interest is $[0, 51.2] {\times} [-25.6, 25.6] {\times} [-2, 4.4]~\mathrm{m}^{3}$, with a voxelization resolution of $0.2~m$, resulting in a voxel grid of $256 {\times} 256 {\times} 32$.
Following prior vision-based methods~\cite{MonoScene,SGN,CGFormer}, we take RGB images as input and predict voxel-wise occupancy and semantic labels within the predefined 3D space.

To assess the applicability of VoxSAOR beyond autonomous driving, we further evaluate it on Human360Occ (H3O)~\cite{shi2026oneocc}, a panoramic 3D occupancy benchmark for embodied perception. Human360Occ~\cite{shi2026oneocc} is a CARLA-based 360{\textdegree} occupancy dataset with a human-centric egocentric viewpoint and simulated gait. 
It comprises $160$ sequences (approximately $8K$ frames) across $16$ maps under diverse weather and time-of-day conditions. 
Each sample spans $[-12.8,12.8]{\times}[-12.8,12.8]{\times}[-2.4,0.8]~\mathrm{m}^3$, with occupancy annotations at native resolutions of $64{\times}64{\times}8$.
The dataset contains $10$ semantic classes plus an \emph{empty} class and supports within-city (\emph{Homo}), using an $8{:}2$ training/validation split per map, and cross-city (\emph{Heter}), using the default $12/4$ map split.

\subsubsection{Metrics}
We mainly report the Intersection-over-Union (IoU) for scene completion and mIoU over the semantic classes (no \emph{``empty''} class) for semantic scene understanding, supplemented by tail mIoU~\cite{zhang2026proood} to assess performance on underrepresented classes. 
Moreover, following~\cite{cao2024pasco,laurent2022packed}, we use the maximum softmax probability as a measure of model confidence. 
Consistent with established practices~\cite{cao2024pasco,havasi2021training,laurent2022packed,lakshminarayanan2017simple}, we assess predictive uncertainty from a calibration perspective~\cite{guo2017calibration} using the Expected Calibration Error (ECE) and Negative Log-Likelihood (NLL).
To account for the dominance of empty voxels in 3D scenes, we follow prior work~\cite{cao2024pasco} to compute voxel-level uncertainty separately for empty and non-empty voxels and report their average.

\begin{table*}[!t]
\newcommand{\clsname}[2]{
\rotatebox{90}{
\hspace{-6pt}
\textcolor{#2}{$\blacksquare$}
\hspace{-6pt}
\renewcommand\arraystretch{0.6}
\begin{tabular}{l}
#1                                      \\
\hspace{-4pt} ~\tiny(\semkitfreq{#2}\%) \\
\end{tabular}
}}
\centering
\caption{\textbf{Quantitative results on SemanticKITTI validation set}~\cite{SemanticKITTI}\textbf{.}
\textbf{T} indicates using extra temporal information. 
The best and the second best results are in  {bold} and  {underlined}, respectively.
}
\label{tab:sem_kitti_val}
\resizebox{\textwidth}{!}{
\begin{tabular}{l|c|cc|*{19}{c}}
    \toprule
    Method & \textbf{T} & IoU & mIoU & \clsname{road}{road} & \clsname{sidewalk}{sidewalk} & \clsname{parking}{parking} & \clsname{other-grnd.}{otherground} & \clsname{building}{building} & \clsname{car}{car} & \clsname{truck}{truck} & \clsname{bicycle}{bicycle} & \clsname{motorcycle}{motorcycle} & \clsname{other-veh.}{othervehicle} & \clsname{vegetation}{vegetation} & \clsname{trunk}{trunk} & \clsname{terrain}{terrain} & \clsname{person}{person} & \clsname{bicyclist}{bicyclist} & \clsname{motorcyclist}{motorcyclist} & \clsname{fence}{fence} & \clsname{pole}{pole} & \clsname{traf.-sign}{trafficsign} \\
    \midrule
    MonoScene~\cite{MonoScene} & & 36.86 & 11.08 & 56.52 & 26.72 & 14.27 & 0.46 & 14.09 & 23.26 & 6.98 & 0.61 & 0.45 & 1.48 & 17.89 & 2.81 & 29.64 & 1.86 & 1.20 & 0.00 & 5.84 & 4.14 & 2.25 \\
    TPVFormer~\cite{TPVFormer} & & 35.61 & 11.36 & 56.50 & 25.87 & 20.60 & 0.85 & 13.88 & 23.81 & 8.08 & 0.36 & 0.05 & 4.35 & 16.92 & 2.26 & 30.38 & 0.51 & 0.89 & 0.00 & 5.94 & 3.14 & 1.52 \\
    OccFormer~\cite{OccFormer} & & 36.50 & 13.46 & {58.85} & 26.88 & 19.61 & 0.31 & 14.40 & 25.09 & {25.53} & 0.81 & 1.19 & {8.52} & 19.63 & 3.93 & 32.62 & 2.78 & 2.82 & 0.00 & 5.61 & 4.26 & 2.86 \\
    IAMSSC~\cite{IAMSSC} & & 44.29 & 12.45 & 54.55 & 25.85 & 16.02 & 0.70 & 17.38 & 26.26 & 8.74 & 0.60 & 0.15 & 5.06 & 24.63 & 4.95 & 30.13 & 1.32 & 3.46 & 0.01 & 6.86 & 6.35 & 3.56 \\
    VoxFormer~\cite{VoxFormer} & & 44.02 & 12.35 & 54.76 & 26.35 & 15.50 & 0.70 & 17.65 & 25.79 & 5.63 & 0.59 & 0.51 & 3.77 & 24.39 & 5.08 & 29.96 & 1.78 & 3.32 & 0.00 & 7.64 & 7.11 & 4.18 \\
    VoxFormer~\cite{VoxFormer} & \checkmark & 44.15 & 13.35 & 53.57 & 26.52 & 19.69 & 0.42 & 19.54 & 26.54 & 7.26 & 1.28 & 0.56 & 7.81 & 26.10 & 6.10 & 33.06 & 1.93 & 1.97 & 0.00 & 7.31 & 9.15 & 4.94 \\
    Symphonies~\cite{Symphonize} & & 41.92 & {14.89} & 56.37 & 27.58 & 15.28 & \SecondBest{0.95} & 21.64 & 28.68 & {20.44} & {2.54} & {2.82} & {13.89} & 25.72 & 6.60 & 30.87 & {3.52} & 2.24 & 0.00 & 8.40 & 9.57 & 5.76 \\
    HASSC~\cite{HASSC} & & 44.82 & 13.48 & 57.05 & 28.25 & 15.90 & \Best{1.04} & 19.05 & 27.23 & 9.91 & 0.92 & 0.86 & 5.61 & 25.48 & 6.15 & 32.94 & {2.80} & \SecondBest{4.71} & 0.00 & 6.58 & 7.68 & 4.05 \\
    H2GFormer~\cite{H2GFormer} & & 44.57 & 13.73 & 56.08 & 29.12 & 17.83 & 0.45 & 19.74 & 28.21 & 10.00 & 0.50 & 0.47 & 7.39 & 26.25 & 6.80 & 34.42 & 1.54 & 2.88 & 0.00 & 7.24 & 7.88 & 4.68 \\
    H2GFormer~\cite{H2GFormer} & \checkmark & 44.69 & 14.29 & 57.00 & {29.37} & {21.74} & 0.34 & 20.51 & 28.21 & 6.80 & 0.95 & 0.91 & {9.32} & {27.44} & 7.80 & {36.26} & 1.15 & 0.10 & 0.00 & 7.98 & {9.88} & 5.81 \\
    SGN~\cite{SGN} & \checkmark & {46.21} & 15.32 & 59.10 & 29.41 & 19.05 & 0.33 & {25.17} & 33.31 & 6.03 & 0.61 & 0.46 & 9.84 & \SecondBest{28.93} & \SecondBest{9.58} & 38.12 & 0.47 & 0.10 & 0.00 & 9.96 & \Best{13.25} & 7.32 \\
    HTCL~\cite{li2024hierarchical} & \checkmark & 45.51 & {17.13} & 63.70 & {32.48} & \SecondBest{23.27} & 0.14 & 24.13 & 34.30 & 20.72 & {3.99} & 2.80 & 11.99 & 26.96 & 8.79 & 37.73 & 2.56 & 2.30 & 0.00 & \SecondBest{11.22} & {11.49} & 6.95 \\
    
    \midrule
    SGN~\cite{SGN} & & 43.60 & 14.55 & 59.32 & 30.51 & 18.46 & 0.42 & 21.43 & 31.88 & 13.18 & 0.58 & 0.17 & 5.68 & 25.98 & 7.43 & 34.42 & 1.28 & 1.49 & 0.00 & 9.66 & 9.83 & 4.71 \\
    {\cellcolor{mygreen!10}}VoxSAOR (+SGN) & {\cellcolor{mygreen!10}} 
    & {\cellcolor{mygreen!10}}44.08 
    & {\cellcolor{mygreen!10}}14.79 
    & {\cellcolor{mygreen!10}}59.62 
    & {\cellcolor{mygreen!10}}31.15 
    & {\cellcolor{mygreen!10}}20.49 
    & {\cellcolor{mygreen!10}}0.15 
    & {\cellcolor{mygreen!10}}21.38 
    & {\cellcolor{mygreen!10}}32.09 
    & {\cellcolor{mygreen!10}}11.07 
    & {\cellcolor{mygreen!10}}0.60 
    & {\cellcolor{mygreen!10}}0.83 
    & {\cellcolor{mygreen!10}}3.31 
    & {\cellcolor{mygreen!10}}26.13 
    & {\cellcolor{mygreen!10}}7.76 
    & {\cellcolor{mygreen!10}}35.41 
    & {\cellcolor{mygreen!10}}2.87 
    & {\cellcolor{mygreen!10}}2.79 
    & {\cellcolor{mygreen!10}}0.00 
    & {\cellcolor{mygreen!10}}8.69 
    & {\cellcolor{mygreen!10}}11.06 
    & {\cellcolor{mygreen!10}}5.70 \\
    
    \midrule
    CGFormer~\cite{CGFormer} & & {45.99} & {16.87} & 65.51 & {32.31} & 20.82 & 0.16 & {23.52} & {34.32} & 19.44 & \SecondBest{4.61} & {2.71} & 7.67 & {26.93} & {8.83} & {39.54} & 2.38 & {4.08} & 0.00 & {9.20} & {10.67} & {7.84} \\
    {\cellcolor{mygreen!10}}VoxSAOR (+CGFormer) & {\cellcolor{mygreen!10}} 
    & {\cellcolor{mygreen!10}}45.64 
    & {\cellcolor{mygreen!10}}17.58 
    & {\cellcolor{mygreen!10}}\Best{66.68} 
    & {\cellcolor{mygreen!10}}34.03 
    & {\cellcolor{mygreen!10}}\Best{23.30} 
    & {\cellcolor{mygreen!10}}0.18 
    & {\cellcolor{mygreen!10}}23.17 
    & {\cellcolor{mygreen!10}}33.81 
    & {\cellcolor{mygreen!10}}22.17 
    & {\cellcolor{mygreen!10}}\Best{4.88} 
    & {\cellcolor{mygreen!10}}\Best{5.96} 
    & {\cellcolor{mygreen!10}}9.75 
    & {\cellcolor{mygreen!10}}27.19 
    & {\cellcolor{mygreen!10}}8.33 
    & {\cellcolor{mygreen!10}}39.45 
    & {\cellcolor{mygreen!10}}2.19 
    & {\cellcolor{mygreen!10}}4.52 
    & {\cellcolor{mygreen!10}}0.00 
    & {\cellcolor{mygreen!10}}11.21 
    & {\cellcolor{mygreen!10}}10.49 
    & {\cellcolor{mygreen!10}}6.74 \\
    
    \midrule
    VoxDet~\cite{VoxDet} & & \SecondBest{47.36} & \SecondBest{18.73} & \SecondBest{65.55} & \Best{34.22} & {20.88} & 0.04 & \SecondBest{25.79} & \Best{34.50} & \SecondBest{31.05} & 3.95 & {5.14} & \SecondBest{14.65} & {28.93} & \Best{10.20} & \SecondBest{41.27} & \SecondBest{4.48} & {3.14} & {0.00} & \Best{11.73} & {12.19} & \SecondBest{8.28} \\
    {\cellcolor{mygreen!10}}VoxSAOR (+VoxDet) & {\cellcolor{mygreen!10}} 
    & {\cellcolor{mygreen!10}}\Best{47.70}
    & {\cellcolor{mygreen!10}}\Best{19.19}
    & {\cellcolor{mygreen!10}}65.08
    & {\cellcolor{mygreen!10}}\SecondBest{34.10}
    & {\cellcolor{mygreen!10}}22.84
    & {\cellcolor{mygreen!10}}0.03
    & {\cellcolor{mygreen!10}}\Best{26.18}
    & {\cellcolor{mygreen!10}}\SecondBest{34.32}
    & {\cellcolor{mygreen!10}}\Best{31.81}
    & {\cellcolor{mygreen!10}}4.05
    & {\cellcolor{mygreen!10}}\SecondBest{5.93}
    & {\cellcolor{mygreen!10}}\Best{15.46}
    & {\cellcolor{mygreen!10}}\Best{29.73}
    & {\cellcolor{mygreen!10}}9.73
    & {\cellcolor{mygreen!10}}\Best{41.80}
    & {\cellcolor{mygreen!10}}\Best{5.72}
    & {\cellcolor{mygreen!10}}\Best{5.32}
    & {\cellcolor{mygreen!10}}0.00
    & {\cellcolor{mygreen!10}}10.97
    & {\cellcolor{mygreen!10}}\SecondBest{12.75}
    & {\cellcolor{mygreen!10}}\Best{8.70} \\
    \bottomrule
\end{tabular}}
\end{table*}

\begin{table*}[!t]
\newcommand{\clsname}[2]{
\rotatebox{90}{
\hspace{-6pt}
\textcolor{#2}{$\blacksquare$}
\hspace{-6pt}
\renewcommand\arraystretch{0.6}
\begin{tabular}{l}
#1                                       \\
\hspace{-4pt} ~\tiny(\sscbkitfreq{#2}\%) \\
\end{tabular}
}}
\centering
\caption{\textbf{Quantitative results on SSCBench-KITTI360 test set}~\cite{li2024sscbench}\textbf{.} 
The best and second best results are in \Best{bold} and \SecondBest{underlined}, respectively. $\ast$ denotes results reproduced using the official code implementation.
}
\label{tab:kitti_360}
\resizebox{\textwidth}{!}{
\begin{tabular}{l|cc|*{18}{c}}
    \toprule
    Method & IoU & mIoU & \clsname{car}{car} & \clsname{bicycle}{bicycle} & \clsname{motorcycle}{motorcycle} & \clsname{truck}{truck} & \clsname{other-veh.}{othervehicle} & \clsname{person}{person} & \clsname{road}{road} & \clsname{parking}{parking} & \clsname{sidewalk}{sidewalk} & \clsname{other-grnd.}{otherground} & \clsname{building}{building} & \clsname{fence}{fence} & \clsname{vegetation}{vegetation} & \clsname{terrain}{terrain} & \clsname{pole}{pole} & \clsname{traf.-sign}{trafficsign} & \clsname{other-struct.}{otherstructure} & \clsname{other-obj.}{otherobject} \\
    \midrule
    \multicolumn{21}{l}{\textit{LiDAR-based methods}} \\
    \midrule
    \textcolor{gray}{SSCNet~\cite{SSCNet}} & \textcolor{gray}{{53.58}} & \textcolor{gray}{16.95} & \textcolor{gray}{{31.95}} & \textcolor{gray}{0.00} & \textcolor{gray}{0.17} & \textcolor{gray}{10.29} & \textcolor{gray}{0.00} & \textcolor{gray}{0.07} & \textcolor{gray}{{65.70}} & \textcolor{gray}{{17.33}} & \textcolor{gray}{{41.24}} & \textcolor{gray}{3.22} & \textcolor{gray}{{44.41}} & \textcolor{gray}{6.77} & \textcolor{gray}{{43.72}} & \textcolor{gray}{{28.87}} & \textcolor{gray}{0.78} & \textcolor{gray}{0.75} & \textcolor{gray}{8.69} & \textcolor{gray}{0.67} \\
    \textcolor{gray}{LMSCNet~\cite{LMSCNet}} & \textcolor{gray}{47.35} & \textcolor{gray}{13.65} & \textcolor{gray}{20.91} & \textcolor{gray}{0.00} & \textcolor{gray}{0.00} & \textcolor{gray}{0.26} & \textcolor{gray}{0.58} & \textcolor{gray}{0.00} & \textcolor{gray}{62.95} & \textcolor{gray}{13.51} & \textcolor{gray}{33.51} & \textcolor{gray}{0.20} & \textcolor{gray}{43.67} & \textcolor{gray}{0.33} & \textcolor{gray}{40.01} & \textcolor{gray}{26.80} & \textcolor{gray}{0.00} & \textcolor{gray}{0.00} & \textcolor{gray}{3.63} & \textcolor{gray}{0.00} \\
    \midrule
    \multicolumn{21}{l}{\textit{Camera-based methods}} \\
    \midrule
    MonoScene~\cite{MonoScene} & 37.87 & 12.31 & 19.34 & 0.43 & 0.58 & 8.02 & 2.03 & 0.86 & 48.35 & 11.38 & 28.13 & 3.32 & 32.89 & 3.53 & 26.15 & 16.75 & 6.92 & 5.67 & 4.20 & 3.09 \\
    TPVFormer~\cite{TPVFormer} & 40.22 & 13.64 & 21.56 & 1.09 & 1.37 & 8.06 & 2.57 & 2.38 & 52.99 & 11.99 & 31.07 & 3.78 & 34.83 & 4.80 & 30.08 & 17.52 & 7.46 & 5.86 & 5.48 & 2.70 \\
    OccFormer~\cite{OccFormer} & 40.27 & 13.81 & 22.58 & 0.66 & 0.26 & 9.89 & 3.82 & 2.77 & 54.30 & 13.44 & 31.53 & 3.55 & 36.42 & 4.80 & 31.00 & 19.51 & 7.77 & 8.51 & 6.95 & 4.60 \\
    VoxFormer~\cite{VoxFormer} & 38.76 & 11.91 & 17.84 & 1.16 & 0.89 & 4.56 & 2.06 & 1.63 & 47.01 & 9.67 & 27.21 & 2.89 & 31.18 & 4.97 & 28.99 & 14.69 & 6.51 & 6.92 & 3.79 & 2.43 \\
    IAMSSC~\cite{IAMSSC} & 41.80 & 12.97 & 18.53 & {2.45} & 1.76 & 5.12 & 3.92 & 3.09 & 47.55 & 10.56 & 28.35 & 4.12 & 31.53 & 6.28 & 29.17 & 15.24 & 8.29 & 7.01 & 6.35 & 4.19 \\
    DepthSSC~\cite{DepthSSC} & 40.85 & 14.28 & 21.90 & 2.36 & {4.30} & 11.51 & 4.56 & 2.92 & 50.88 & 12.89 & 30.27 & 2.49 & {37.33} & 5.22 & 29.61 & {21.59} & 5.97 & 7.71 & 5.24 & 3.51 \\
    Symphonies~\cite{Symphonize} & {43.41} & {17.82} & {26.86} & {4.21} & {4.90} & {14.20} & {7.76} & {6.57} & {57.30} & {13.58} & {35.24} & {4.57} & 39.20 & {7.95} & {34.33} & 19.19 & {14.04} & {15.78} & {8.23} & {6.04} \\
    SGN\text{-}S~\cite{SGN} & 46.22 & 17.71 & 28.20 & 2.09 & 3.02 & 11.95 & 3.68 & 4.20 & 59.49 & 14.50 & 36.53 & 4.24 & 39.79 & 7.14 & 36.61 & 23.10 & 14.86 & 16.14 & 8.24 & 4.95 \\
    SGN\text{-}T~\cite{SGN} & 47.06 & 18.25 & 29.03 & 3.43 & 2.90 & 10.89 & 5.20 & 2.99 & 58.14 & 15.04 & 36.40 & 4.43 & 42.02 & 7.72 & 38.17 & 23.22 & {16.73} & 16.38 & 9.93 & 5.86 \\
    SGFormer~\cite{guo2025sgformer} & 46.35 & 18.30 & 27.80 & 0.91 & 2.55 & 10.73 & 5.67 & 4.28 & 61.04 & 13.21 & 37.00 & 5.07 & {43.05} & 7.46 & {38.98} & 24.87 & 15.75 & 16.90 & 8.85 & 5.33 \\

    \midrule
    CGFormer~\cite{CGFormer} & {48.07} & {20.05} & \SecondBest{29.85} & {3.42} & 3.96 & \Best{17.59} & {6.79} & {6.63} & \Best{63.85} & {17.15} & \SecondBest{40.72} & {5.53} & {42.73} & {8.22} & {38.80} & \SecondBest{24.94} & {16.24} & {17.45} & {10.18} & {6.77} \\
    {\cellcolor{mygreen!10}}VoxSAOR (+CGFormer) 
    & {\cellcolor{mygreen!10}}48.03 
    & {\cellcolor{mygreen!10}}20.40 
    & {\cellcolor{mygreen!10}}29.84 
    & {\cellcolor{mygreen!10}}3.61 
    & {\cellcolor{mygreen!10}}4.43 
    & {\cellcolor{mygreen!10}}16.35 
    & {\cellcolor{mygreen!10}}\Best{8.62} 
    & {\cellcolor{mygreen!10}}6.72 
    & {\cellcolor{mygreen!10}}\SecondBest{63.62} 
    & {\cellcolor{mygreen!10}}\Best{18.50} 
    & {\cellcolor{mygreen!10}}\Best{41.18} 
    & {\cellcolor{mygreen!10}}\SecondBest{5.88} 
    & {\cellcolor{mygreen!10}}42.91 
    & {\cellcolor{mygreen!10}}9.00 
    & {\cellcolor{mygreen!10}}\Best{39.07} 
    & {\cellcolor{mygreen!10}}\Best{25.34} 
    & {\cellcolor{mygreen!10}}15.88 
    & {\cellcolor{mygreen!10}}18.86 
    & {\cellcolor{mygreen!10}}10.22 
    & {\cellcolor{mygreen!10}}7.17 \\
    
    \midrule
    VoxDet$^\ast$~\cite{VoxDet} & \SecondBest{48.42} & \SecondBest{20.72} & 29.57 & \SecondBest{5.04} & \SecondBest{6.56} & 16.35 & 8.36 & \SecondBest{7.49} & 62.24 & 17.59 & 40.52 & \Best{5.93} & \Best{44.09} & \Best{9.87} & 38.78 & 23.51 & \SecondBest{17.48} & \SecondBest{19.95} & \SecondBest{11.37} & \SecondBest{8.20} \\
    {\cellcolor{mygreen!10}}VoxSAOR (+VoxDet) 
    & {\cellcolor{mygreen!10}}\Best{48.73} 
    & {\cellcolor{mygreen!10}}\Best{21.06} 
    & {\cellcolor{mygreen!10}}\Best{29.86} 
    & {\cellcolor{mygreen!10}}\Best{5.23} 
    & {\cellcolor{mygreen!10}}\Best{8.46} 
    & {\cellcolor{mygreen!10}}\SecondBest{16.97} 
    & {\cellcolor{mygreen!10}}\SecondBest{8.41} 
    & {\cellcolor{mygreen!10}}\Best{8.06} 
    & {\cellcolor{mygreen!10}}62.47 
    & {\cellcolor{mygreen!10}}\SecondBest{18.28} 
    & {\cellcolor{mygreen!10}}40.42 
    & {\cellcolor{mygreen!10}}5.33 
    & {\cellcolor{mygreen!10}}\SecondBest{43.98} 
    & {\cellcolor{mygreen!10}}\SecondBest{9.73} 
    & {\cellcolor{mygreen!10}}\SecondBest{39.03} 
    & {\cellcolor{mygreen!10}}24.41 
    & {\cellcolor{mygreen!10}}\Best{17.49} 
    & {\cellcolor{mygreen!10}}\Best{20.48} 
    & {\cellcolor{mygreen!10}}\Best{11.66} 
    & {\cellcolor{mygreen!10}}\Best{8.75} \\
    \bottomrule
\end{tabular}}
\end{table*}

\subsection{Implementation Details}
We implement VoxSAOR on existing vision-based 3D occupancy prediction backbones~\cite{VoxDet,MonoScene,SGN,CGFormer}, retaining their default training configurations, including the optimizer, learning-rate schedule, and backbone-specific loss settings. For cross-view semantic consensus, we use OpenSeeD~\cite{zhang2023simple} to generate image-space semantic pseudo-labels offline. 
These pseudo-labels are used only during training to filter unreliable semantic correspondences and are not required at inference.
For LSC, we adopt the full $26$-connected neighborhood, with connectivity weights $\lambda_{\mathrm{face}}{=}1$, $\lambda_{\mathrm{edge}}{=}0.1$, and $\lambda_{\mathrm{corner}}{=}0.3$.
For stable-to-ambiguous transition construction, we set the number of ambiguous anchors to $n_a{=}4$ and the minimum number of candidate voxels per semantic category to $n_{\min}{=}32$. 
For AREL, we use $R{=}2$ prediction regimes with a base scale of $\alpha{=}0.01$, where $\alpha_m{=}m\alpha$ for $m{=}1,\ldots,R$.
Each branch is supervised using the original occupancy objective of the backbone, and the head losses are averaged for joint optimization. 
During inference, SAUM and proxy sampling are disabled, and the probability predictions of all heads are averaged to obtain the final occupancy prediction.

\subsection{Main Results}

\subsubsection{3D Occupancy Prediction}

\begin{table*}[!t]
    \centering
    \caption{\textbf{Occupancy prediction results on H3O (Human360Occ)}~\cite{shi2026oneocc}\textbf{.}
    Two vertically stacked panels report results on \textbf{HOMO (within-city)} and \textbf{HETER (cross-city)} splits.
    }
    \label{tab:h3o_homo_heter}

    {\small
    \resizebox{.8\textwidth}{!}{
    \begin{tabular}{l|cc|cccccccccc}
        \toprule
        \rowcolor{gray!20}
        \multicolumn{13}{c}{\emph{\textbf{HOMO (within-city)}}}\\
        \midrule
        Method & IoU & mIoU
        & \ColHeadHomo{hthreeoRoad}{road}{road}
        & \ColHeadHomo{hthreeoSidewalk}{sidewalk}{sidewalk}
        & \ColHeadHomo{hthreeoBuilding}{building}{building}
        & \ColHeadHomo{hthreeoVegetation}{vegetation}{vegetation}
        & \ColHeadHomo{hthreeoCar}{car}{car}
        & \ColHeadHomo{hthreeoTruck}{truck}{truck}
        & \ColHeadHomo{hthreeoBus}{bus}{bus}
        & \ColHeadHomo{hthreeoTwoWheeler}{two\_wheeler}{twoWheeler}
        & \ColHeadHomo{hthreeoPerson}{person}{person}
        & \ColHeadHomo{hthreeoPole}{pole}{pole} \\
        \midrule
        VoxFormer\text{-}S~\cite{VoxFormer} & 31.57 & 11.09 & 22.27 & 17.44 & 2.86 & 5.65 & 0.87 & 0.14 & 0.00 & 0.00 & 61.46 & 0.18 \\
        SGN\text{-}T~\cite{SGN} & 46.05 & 28.64 & 54.55 & 44.61 & 15.51 & 35.16 & 23.91 & 7.68 & 4.52 & 15.77 & 65.24 & 19.46 \\
        SGN\text{-}S~\cite{SGN} & 45.98 & 28.56 & 54.81 & 47.44 & 15.64 & 33.22 & 24.23 & 8.53 & 2.62 & 13.51 & 65.88 & 19.72 \\
        OccFormer~\cite{OccFormer} & 45.98 & 24.88 & 54.11 & 46.41 & 15.14 & 32.84 & 17.57 & 4.30 & 0.00 & 1.65 & 65.08 & 11.72 \\
        OneOcc~\cite{shi2026oneocc} & \SecondBest{55.73} & \Best{37.29} & \Best{63.82} & \Best{62.92} & \Best{22.86} & \SecondBest{41.03} & \Best{33.74} & \Best{17.79} & \SecondBest{6.47} & \Best{26.58} & \Best{71.13} & \SecondBest{26.55} \\
        \midrule
        MonoScene~\cite{MonoScene} & 54.08 & 33.46 & \SecondBest{62.45} & 57.18 & 19.97 & 38.91 & 29.76 & 13.77 & 5.03 & 16.76 & 70.15 & 20.67 \\
        {\cellcolor{mygreen!10}}VoxSAOR (+MonoScene) 
        & {\cellcolor{mygreen!10}}\Best{55.99}
        & {\cellcolor{mygreen!10}}\SecondBest{36.40}
        & {\cellcolor{mygreen!10}}62.32
        & {\cellcolor{mygreen!10}}\SecondBest{57.75}
        & {\cellcolor{mygreen!10}}\SecondBest{21.77}
        & {\cellcolor{mygreen!10}}\Best{43.45}
        & {\cellcolor{mygreen!10}}\SecondBest{32.99}
        & {\cellcolor{mygreen!10}}\SecondBest{15.80}
        & {\cellcolor{mygreen!10}}\Best{10.84}
        & {\cellcolor{mygreen!10}}\SecondBest{21.05}
        & {\cellcolor{mygreen!10}}\SecondBest{70.48}
        & {\cellcolor{mygreen!10}}\Best{27.56} \\
        \bottomrule
    \end{tabular}
    }}

    \vspace{2ex}

    {\small
    \resizebox{.8\textwidth}{!}{
    \begin{tabular}{l|cc|cccccccccc}
        \toprule
        \rowcolor{gray!20}
        \multicolumn{13}{c}{\emph{\textbf{HETER (cross-city)}}}\\
        \midrule
        Method & IoU & mIoU
        & \ColHeadHeter{hthreeoRoad}{road}{road}
        & \ColHeadHeter{hthreeoSidewalk}{sidewalk}{sidewalk}
        & \ColHeadHeter{hthreeoBuilding}{building}{building}
        & \ColHeadHeter{hthreeoVegetation}{vegetation}{vegetation}
        & \ColHeadHeter{hthreeoCar}{car}{car}
        & \ColHeadHeter{hthreeoTruck}{truck}{truck}
        & \ColHeadHeter{hthreeoBus}{bus}{bus}
        & \ColHeadHeter{hthreeoTwoWheeler}{two\_wheeler}{twoWheeler}
        & \ColHeadHeter{hthreeoPerson}{person}{person}
        & \ColHeadHeter{hthreeoPole}{pole}{pole} \\
        \midrule
        VoxFormer\text{-}S~\cite{VoxFormer} & 27.40 & 10.63 & 19.02 & 15.82 & 2.84 & 5.86 & 0.95 & 0.10 & 0.00 & 0.01 & 61.62 & 0.10 \\
        SGN\text{-}T~\cite{SGN} & 30.26 & 20.02 & 42.57 & 23.94 & 9.42 & 15.84 & 18.22 & 10.21 & 1.55 & 3.11 & 63.44 & 11.93 \\
        SGN\text{-}S~\cite{SGN} & 32.76 & 20.60 & 43.47 & 24.70 & 9.19 & 20.27 & 17.21 & 8.21 & 1.85 & 5.08 & 61.10 & 14.89 \\
        OccFormer~\cite{OccFormer} & 40.73 & 20.87 & 47.00 & 39.01 & 10.14 & 24.85 & 11.84 & 1.52 & 0.05 & 0.77 & 66.60 & 6.93 \\
        OneOcc~\cite{shi2026oneocc} & \Best{49.58} & \Best{32.23} & \Best{58.60} & \Best{48.28} & \Best{16.34} & \Best{25.98} & \Best{30.68} & \Best{19.35} & \Best{12.87} & \Best{14.16} & \Best{72.89} & \Best{23.11} \\
        \midrule
        MonoScene~\cite{MonoScene} & 43.44 & 24.15 & 49.68 & \SecondBest{41.41} & 11.04 & 20.57 & 20.34 & 10.36 & 1.98 & 2.79 & \SecondBest{69.68} & 13.69 \\
        {\cellcolor{mygreen!10}}VoxSAOR (+MonoScene) 
        & {\cellcolor{mygreen!10}}\SecondBest{44.47}
        & {\cellcolor{mygreen!10}}\SecondBest{28.52}
        & {\cellcolor{mygreen!10}}\SecondBest{52.00}
        & {\cellcolor{mygreen!10}}40.00
        & {\cellcolor{mygreen!10}}\SecondBest{13.27}
        & {\cellcolor{mygreen!10}}\SecondBest{25.29}
        & {\cellcolor{mygreen!10}}\SecondBest{25.19}
        & {\cellcolor{mygreen!10}}\SecondBest{18.36}
        & {\cellcolor{mygreen!10}}\SecondBest{11.60}
        & {\cellcolor{mygreen!10}}\SecondBest{9.00}
        & {\cellcolor{mygreen!10}}69.50
        & {\cellcolor{mygreen!10}}\SecondBest{20.98} \\
        \bottomrule
    \end{tabular}
    }}
\end{table*}

Tables~\ref{tab:sem_kitti_test} and~\ref{tab:sem_kitti_val} present the quantitative results on the SemanticKITTI hidden test and validation sets~\cite{SemanticKITTI}, respectively.
VoxSAOR consistently improves mIoU across SGN~\cite{SGN}, CGFormer~\cite{CGFormer}, and VoxDet~\cite{VoxDet}, yielding gains of $0.36$, $0.42$, and $0.41$ percentage points on the test set and $0.24$, $0.71$, and $0.46$ points on the validation set, respectively. Notably, integrating VoxSAOR into VoxDet~\cite{VoxDet} increases IoU from $46.88\%$ to $48.02\%$ and mIoU from $18.23\%$ to $18.64\%$ on the test set. These consistent mIoU improvements highlight the effectiveness of VoxSAOR in enhancing semantic discrimination across different occupancy architectures, aligning with the objective of semantic-aware occupancy robustification to improve semantic predictions under observation-induced feature variations.

\begin{figure*}[!t]
    \centering
    \includegraphics[width=\textwidth]{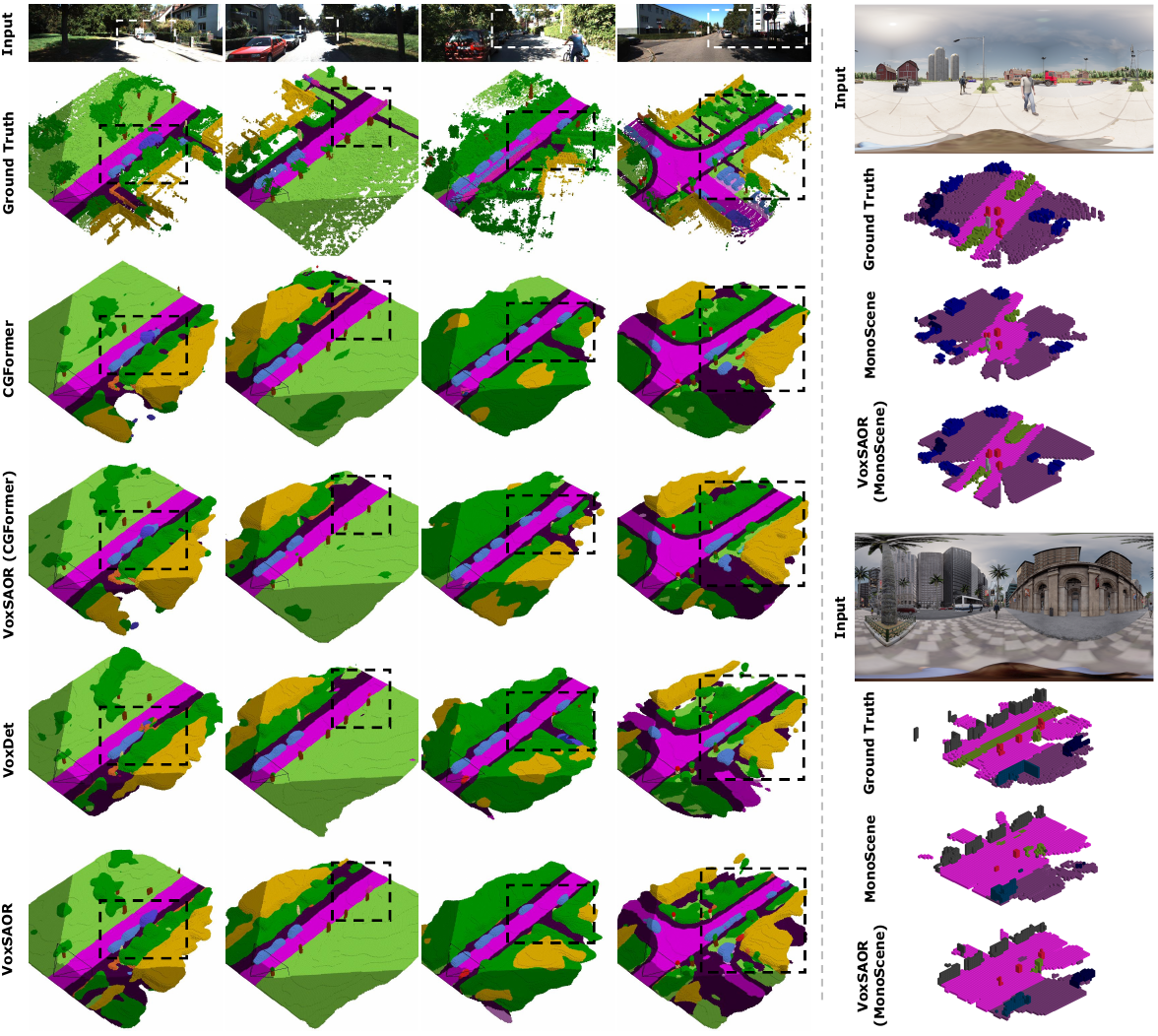}
    \caption{\textbf{Qualitative comparison of 3D occupancy prediction} on SemanticKITTI validation set~\cite{SemanticKITTI} (left) and Human360Occ~\cite{shi2026oneocc} (right). 
    On SemanticKITTI, CGFormer~\cite{CGFormer} and VoxDet~\cite{VoxDet} are compared against VoxSAOR, while MonoScene~\cite{MonoScene} serves as the backbone on Human360Occ. Dashed boxes highlight challenging regions for structural comparison.}
    \label{fig:occ_vis}
\end{figure*}

To further assess its applicability across perception scenarios, we evaluate VoxSAOR on SSCBench-KITTI360~\cite{li2024sscbench} and Human360Occ~\cite{shi2026oneocc}. As shown in Table~\ref{tab:kitti_360}, VoxSAOR improves the mIoU of CGFormer~\cite{CGFormer} and VoxDet~\cite{VoxDet} by $0.35$ and $0.34$ percentage points, respectively, while also yielding an IoU gain of $0.31$ points for VoxDet. For panoramic embodied perception, Table~\ref{tab:h3o_homo_heter} reports the results on Human360Occ~\cite{shi2026oneocc} under within-city (\emph{Homo}) and cross-city (\emph{Heter}) settings. Using MonoScene~\cite{MonoScene} as the backbone, VoxSAOR improves IoU and mIoU from $54.08\%$ and $33.46\%$ to $55.99\%$ and $36.40\%$, respectively, on the Homo split. On the more challenging Heter split, IoU and mIoU increase from $43.44\%$ and $24.15\%$ to $44.47\%$ and $28.52\%$, corresponding to gains of $1.03$ and $4.37$ percentage points, respectively. Improvements are also observed in several challenging categories, including trucks, buses, and two-wheelers.
Fig.~\ref{fig:occ_vis} further presents qualitative comparisons on SemanticKITTI~\cite{SemanticKITTI} and Human360Occ~\cite{shi2026oneocc}. Compared with the corresponding baselines, VoxSAOR achieves better structural continuity and semantic delineation in challenging regions.
On SemanticKITTI~\cite{SemanticKITTI}, local geometric structures and semantic boundaries are better preserved.
Similarly, on Human360Occ~\cite{shi2026oneocc}, VoxSAOR produces more coherent scene layouts and local semantic structures. These qualitative observations further illustrate the benefits of VoxSAOR across different 3D perception scenarios.

\subsubsection{Uncertainty Estimation and Long-tail Prediction}

\begin{table}[!t]
\centering
\caption{\textbf{Semantic occupancy prediction and predictive calibration on the SemanticKITTI validation set}~\cite{SemanticKITTI}\textbf{.}
Tail mIoU is defined following~\cite{zhang2026proood}.
ECE evaluates predictive calibration, while NLL measures probabilistic prediction quality. Lower ECE and NLL values indicate better performance.}
\label{tab:uncertainty}
\resizebox{\linewidth}{!}{
\begin{tabular}{@{}l|ccc|cc@{}}
    \toprule
    \textbf{Method} & \textbf{IoU} & \textbf{mIoU} & \textbf{Tail mIoU}& \textbf{ECE}~$\downarrow$ & \textbf{NLL}~$\downarrow$ \\
    \midrule
    SGN~\cite{SGN} & 43.60& 14.55& 4.07& 0.2309& 1.5635\\
    VoxSAOR (+SGN) & 44.08& 14.79& 4.19& 0.2301& 1.4749\\
    \midrule
    CGFormer~\cite{CGFormer} & 45.99& 16.87& 6.22& 0.2212& 1.2581\\
    VoxSAOR (+CGFormer) & 45.64& 17.58& 6.84& 0.1972& 0.9619\\
    \midrule
    VoxDet~\cite{VoxDet} & \SecondBest{47.36}& \SecondBest{18.73}& \SecondBest{8.47}& \SecondBest{0.1626}& \SecondBest{0.8397}\\
    VoxSAOR (+VoxDet) & \Best{47.70}& \Best{19.19}& \Best{9.04}& \Best{0.1621}& \Best{0.8298}\\
    \bottomrule
\end{tabular}
}
\end{table}

Beyond semantic occupancy accuracy, we further evaluate predictive calibration and long-tail recognition on the SemanticKITTI validation set~\cite{SemanticKITTI}, as reported in Table~\ref{tab:uncertainty}.
VoxSAOR consistently reduces both Expected Calibration Error (ECE) and Negative Log-Likelihood (NLL) across SGN~\cite{SGN}, CGFormer~\cite{CGFormer}, and VoxDet~\cite{VoxDet}.
For CGFormer~\cite{CGFormer}, ECE decreases from $0.2212$ to $0.1972$, while NLL decreases from $1.2581$ to $0.9619$. For VoxDet~\cite{VoxDet}, the corresponding metrics decrease from $0.1626$ and $0.8397$ to $0.1621$ and $0.8298$, respectively. Although the improvements vary across backbones, these consistent reductions suggest that modeling aleatoric variations in the latent feature space enables the predictors to better account for ambiguity in their semantic decisions, indicating improved confidence calibration and probabilistic prediction quality.
VoxSAOR also consistently improves Tail mIoU, with gains of $0.62$ and $0.57$ percentage points for CGFormer~\cite{CGFormer} and VoxDet~\cite{VoxDet}, respectively. These improvements indicate that the benefits of VoxSAOR extend beyond frequently occurring semantic classes to underrepresented categories. Together with the overall mIoU gains, these results suggest that proxy-guided robustification benefits long-tail recognition alongside overall semantic prediction.

\begin{figure}[!t]
    \centering
    \includegraphics[width=\linewidth]{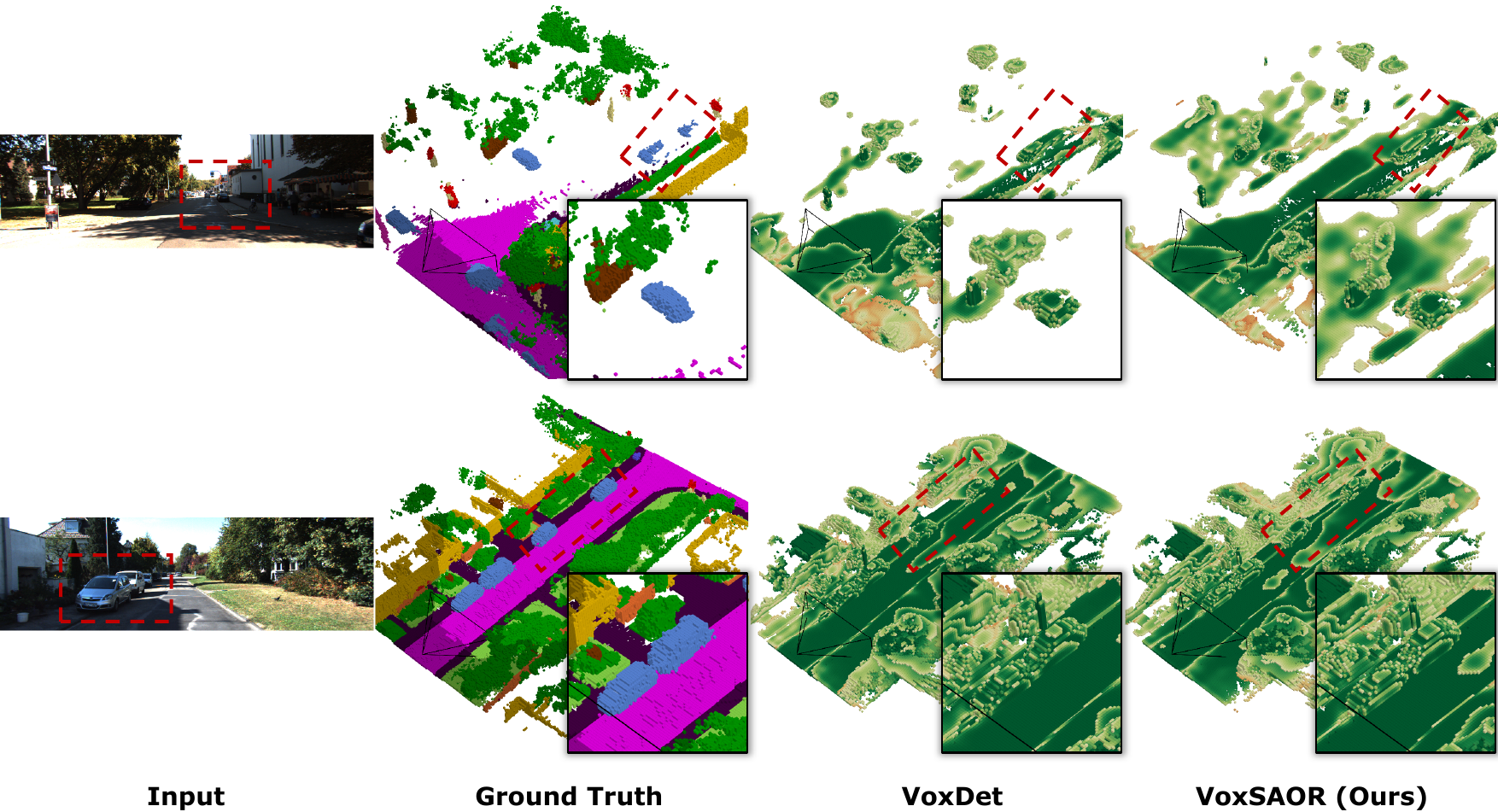}
    \caption{\textbf{Qualitative comparison of uncertainty estimation} on the SemanticKITTI validation set~\cite{SemanticKITTI}. Red dashed boxes highlight challenging regions, with enlarged views illustrating local uncertainty variations. Uncertainty is represented by the complement of the maximum softmax probability, where green and red indicate lower and higher uncertainty, respectively.}
    \label{fig:qualitativeUnc}
\end{figure}

Furthermore, Fig.~\ref{fig:qualitativeUnc} presents qualitative comparisons of voxel-wise uncertainty estimation between VoxDet~\cite{VoxDet} and VoxSAOR on two representative scenes.
The highlighted regions illustrate confidence variations associated with complex object configurations and local geometric structures. 
High-confidence responses are largely preserved in spatially consistent regions, whereas lower confidence appears around structural transitions and object boundaries. Compared with VoxDet~\cite{VoxDet}, VoxSAOR exhibits more spatially differentiated confidence patterns in the highlighted regions while retaining confident predictions over coherent structures.

\subsection{Ablation Studies}
Unless otherwise specified, all ablation studies are conducted on the SemanticKITTI validation set~\cite{SemanticKITTI}, with VoxDet~\cite{VoxDet} as the backbone for VoxSAOR.

\begin{table}[!t]
\centering
\caption{\textbf{Component-wise ablation study.}
SAUM and AREL denote Semantic-Aware Aleatoric Uncertainty Modeling
and Aleatoric Regime Ensemble Learning, respectively.}
\label{tab:components}
\begin{tabular}{@{}cc|cc@{}}
    \toprule
    \textbf{SAUM} & \textbf{AREL}& \textbf{mIoU}& \textbf{IoU} \\
    \midrule
    & & 18.59& 47.51\\
    \ding{51} & & \SecondBest{18.88}& \Best{47.73}\\
    & \ding{51} & 18.53& 47.51\\
    \ding{51} & \ding{51} & \Best{19.19}& \SecondBest{47.70}\\
    \bottomrule
\end{tabular}
\end{table}

\subsubsection{Component Analysis}
Table~\ref{tab:components} evaluates the contributions of two key components of VoxSAOR. Introducing SAUM improves mIoU/IoU from $18.59$/$47.51$ to $18.88$/$47.73$, demonstrating the benefits of semantic-aware aleatoric proxy modeling. 
Incorporating AREL further increases mIoU to $19.19$, a gain of $0.60$ points over the baseline, while maintaining a comparable IoU of $47.70$. 
These results suggest that multi-regime learning further enhances semantic discrimination while largely preserving scene completion performance.

\begin{table}[!t]
\centering
\caption{\textbf{Ablation study of Semantic-Aware Aleatoric Uncertainty Modeling (SAUM).}
Semantic and Spatial denote the same-category constraint and spatial matching used for stable-to-ambiguous transition construction, respectively.}
\label{tab:sau}
\begin{tabular}{@{}c|c|cc|cc@{}}
    \toprule
    \textbf{CSC} & \textbf{LSC} & \textbf{Semantic} & \textbf{Spatial} & \textbf{mIoU} & \textbf{IoU} \\
    \midrule
    \ding{51} & & \ding{51} & \ding{51} & 18.45& \Best{47.71}\\
    \ding{51} & \ding{51} & & & 18.46& 47.63\\
    \ding{51} & \ding{51} & \ding{51} & & \SecondBest{18.88}& 47.51\\
    \ding{51} & \ding{51} & & \ding{51} & 18.62& 47.58\\
    & \ding{51} & \ding{51} & \ding{51} & 18.27& 47.62\\
    \ding{51} & \ding{51} & \ding{51} & \ding{51} & \Best{19.19}& \SecondBest{47.70}\\
    \bottomrule
\end{tabular}
\vskip -2ex
\end{table}

\subsubsection{Ablation of SAUM}
Table~\ref{tab:sau} examines the key designs for constructing the semantic-aware aleatoric proxy.
Removing CSC decreases mIoU from $19.19$ to $18.27$, highlighting the importance of filtering semantically inconsistent cross-view correspondences for reliable transition construction.
Without LSC, mIoU similarly drops to $18.45$, confirming the benefit of structure-driven stratification between stable and ambiguous voxels. We further study the semantic and spatial constraints in transition matching. Removing the semantic or spatial constraint reduces mIoU to $18.62$ and $18.88$, respectively, while discarding both yields only $18.46$ mIoU. 
The complete SAUM configuration achieves the best mIoU of $19.19$ with a comparable IoU of $47.70$, demonstrating that reliable candidate filtering, structural stratification, and constrained transition matching are complementary for constructing an effective aleatoric uncertainty proxy.

\begin{table}[!t]
\centering
\caption{\textbf{Effect of the number of AREL branches.}
\#Params reports the total number of model parameters.}
\label{tab:branches}
\begin{tabular}{@{}c|cc|cc|c@{}}
    \toprule
    \textbf{$R$} & \textbf{mIoU} & \textbf{IoU} & \textbf{ECE} & \textbf{NLL} & \textbf{\#Params}~(\textbf{M}) \\
    \midrule
    1 & \SecondBest{18.88}& \Best{47.73}& 0.1636& 0.8413& 55.34\\
    2 & \Best{19.19}& 47.70& \SecondBest{0.1621}& 0.8298& 55.56\\
    3 & 18.84& \SecondBest{47.71}& \Best{0.1559}& \Best{0.8102}& 55.78\\
    4 & 18.62& 47.59& 0.1623& \SecondBest{0.8267}& 56.01\\
    \bottomrule
\end{tabular}
\end{table}

\subsubsection{Number of Branches}
As shown in Table~\ref{tab:branches}, using two AREL branches achieves the highest mIoU of $19.19$, outperforming the single-branch configuration by $0.31$ percentage points with only $0.22$M additional parameters.
Increasing the number of branches to three further improves uncertainty calibration, achieving the lowest ECE of $0.1559$ and NLL of $0.8102$, but reduces mIoU to $18.84$. A fourth branch brings no further benefit and instead decreases mIoU to $18.62$. Meanwhile, scene-completion IoU remains relatively stable across different branch configurations, suggesting that the branch number has a more pronounced influence on semantic accuracy and predictive reliability. We therefore adopt $R{=}2$ as the default setting, achieving the best semantic accuracy while balancing predictive calibration and model complexity.

\begin{table}[!t]
\centering
\caption{\textbf{Effect of the perturbation scale $\alpha$.}}
\label{tab:alpha}
\begin{tabular}{c|cc}
    \toprule
    $\alpha$ & mIoU& IoU\\
    \midrule
    0.005 & 18.60& \SecondBest{47.73}\\
    0.01 & \Best{19.19}& 47.70\\
    0.02 & \SecondBest{18.95}& 47.61\\
    0.03 & 18.79& \Best{48.03}\\
    0.04 & 18.65& 47.35\\
    \bottomrule
\end{tabular}
\end{table}

\subsubsection{Base Perturbation Scale}
Table~\ref{tab:alpha} studies the effect of the base perturbation scale $\alpha$ used for proxy-induced learning. Increasing $\alpha$ from $0.005$ to $0.01$ improves mIoU from $18.60$ to $19.19$, indicating that a moderate perturbation strength provides more effective supervision across plausible aleatoric variations. Further increasing $\alpha$, however, gradually reduces semantic accuracy, with mIoU decreasing to $18.65$ at $\alpha{=}0.04$. This suggests that overly strong perturbations may move features beyond a useful local uncertainty neighborhood and thus weaken semantic discrimination.

\begin{table}[!t]
\centering
\caption{\textbf{Ablation of the distribution statistics in the aleatoric proxy.}
$\boldsymbol{\mu}_k$ and $\boldsymbol{\sigma}_k^2$ denote the class-conditional channel-wise mean and variance, respectively.}
\label{tab:proxy}
\begin{tabular}{@{}cc|cc@{}}
    \toprule
    $\boldsymbol{\mu}_k$ & $\boldsymbol{\sigma}_k^2$ & \textbf{mIoU} & \textbf{IoU} \\
    \midrule
    & & \SecondBest{18.43}& \Best{47.77}\\
    \ding{51} & & 18.38& 47.36\\
    \ding{51} & \ding{51}& \Best{19.19}& \SecondBest{47.70}\\
    \bottomrule
\end{tabular}
\end{table}

\subsubsection{Aleatoric Proxy Distribution}
Table~\ref{tab:proxy} investigates the effect of distribution statistics in class-conditional aleatoric proxy modeling. Using only the mean $\boldsymbol{\mu}_k$, which captures the dominant stable-to-ambiguous transition direction, does not improve performance and slightly decreases mIoU to $18.38$.
In contrast, jointly modeling the mean and channel-wise variance $\boldsymbol{\sigma}_k^2$ substantially improves mIoU to $19.19$, corresponding to gains of $0.76$ and $0.81$ points over the variant that uses a standard normal distribution (without estimated distribution statistics) and the mean only variant, respectively.
Meanwhile, IoU remains comparable at $47.70$. These results indicate that the directional tendency alone is insufficient to characterize aleatoric feature variations, while additionally capturing their stochastic dispersion yields a substantially more effective uncertainty proxy for robust semantic occupancy prediction.

\begin{table}[!t]
\centering
\caption{\textbf{Sensitivity to the anchor-construction hyperparameters $n_a$ and $n_{\min}$.}}
\label{tab:anchor_ablation}
\begin{tabular}{c|cc@{\qquad}|c|cc}
\toprule
$n_a$& \textbf{mIoU}& \textbf{IoU}&
$n_{\min}$& \textbf{mIoU}& \textbf{IoU}\\
\midrule
2& \SecondBest{18.93}& \SecondBest{47.76}&
24& 18.62& 47.42\\
4& \Best{19.19}& 47.70&
28& \SecondBest{18.80}& \SecondBest{47.82}\\
6& 18.83& 47.58&
32& \Best{19.19}& 47.70\\
8& 18.73& 47.71&
36& 18.51& \Best{47.86}\\
10& 18.58&\Best{47.86}&
40& 18.35& 47.45\\
\bottomrule
\end{tabular}
\end{table}

\begin{table*}[!t]
    \centering
    \caption{\textbf{Ablation study of Local Semantic Consistency (LSC) quantification.}}
    \label{tab:lsc_ablation}
    \begin{minipage}[c]{0.35\textwidth}
        \centering
        \subfloat[Effect of neighborhood connectivity.
        \label{tab:connectivity}]{
            \makebox[\linewidth][c]{%
                \renewcommand{\arraystretch}{1.12}
                \setlength{\tabcolsep}{9pt}
                \begin{tabular}{@{}ccc@{}}
                    \toprule
                    $\mathcal{N}$ & \textbf{mIoU} & \textbf{IoU} \\
                    \midrule
                    $\mathcal{N}_{6}$ & 18.78 & \Best{47.80} \\
                    $\mathcal{N}_{18}$ & \SecondBest{18.80} & 47.39 \\
                    $\mathcal{N}_{26}$ & \Best{19.19} & \SecondBest{47.70} \\
                    \bottomrule
                \end{tabular}%
            }
        }
    \end{minipage}
    \begin{minipage}[c]{0.44\textwidth}
        \centering
        \subfloat[Sensitivity to edge and corner weights.
        \label{fig:weight}]{
            \includegraphics[width=\linewidth]{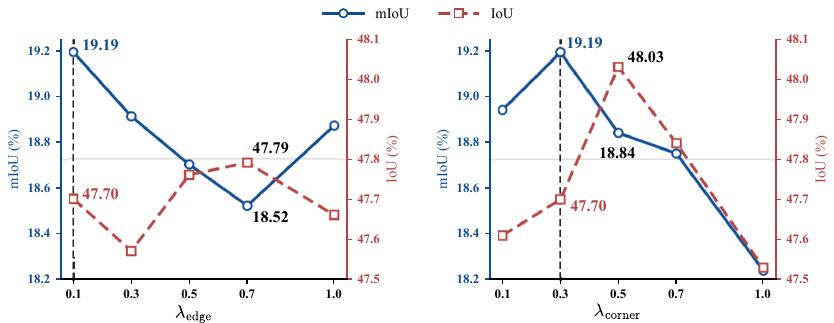}
        }
    \end{minipage}
    \vskip -6ex
\end{table*}

\subsubsection{Anchor Construction}
As indicated in Table~\ref{tab:anchor_ablation}, both $n_a$ and $n_{\min}$ exhibit a clear optimum. 
For the number of ambiguous anchors $n_a$, increasing it from $2$ to $4$ improves mIoU from $18.93$ to $19.19$, whereas using more anchors progressively degrades semantic performance, reaching $18.58$ mIoU at $n_a{=}10$. This suggests that a small set of highly ambiguous voxels provides more representative transition samples, while including too many anchors may introduce less distinctive stable-to-ambiguous variations. 
For $n_{\min}$, the best mIoU of $19.19$ is obtained at $n_{\min}{=}32$.
A smaller threshold may admit categories with insufficient samples for reliable anchor construction, whereas an overly large threshold can exclude too many categories from proxy estimation.
Accordingly, we set $n_a{=}4$ and $n_{\min}{=}32$ by default.

\subsubsection{Local Semantic Consistency Quantification}
Using the full $\mathcal{N}_{26}$ neighborhood achieves the highest mIoU of $19.19$ in Table~\ref{tab:lsc_ablation}, outperforming $\mathcal{N}_6$ and $\mathcal{N}_{18}$ by $0.41$ and $0.39$ points, respectively. This indicates that jointly incorporating face-, edge-, and corner-connected neighbors provides a more informative characterization of the local semantic topology for stable-ambiguous voxel stratification. 
We further evaluate the connectivity-aware weights and obtain the best mIoU with $\lambda_{\mathrm{edge}}{=}0.1$ and $\lambda_{\mathrm{corner}}{=}0.3$. Increasing either weight beyond this setting generally degrades semantic accuracy, suggesting that edge- and corner-connected neighbors are more effective as complementary contextual cues.

\section{Conclusion}
In this work, we presented VoxSAOR, a semantic-aware occupancy robustification framework for vision-based 3D occupancy prediction via aleatoric uncertainty proxy modeling. VoxSAOR constructs class-conditional proxies from stable-to-ambiguous feature transitions to characterize observation-induced aleatoric variations in the voxel feature space. Guided by these proxies, AREL trains multiple prediction heads under distinct perturbation regimes and combines their predictions through deterministic, sampling-free inference. Extensive experiments across multiple datasets and occupancy backbones demonstrate improvements in semantic occupancy prediction, long-tail recognition, and predictive calibration, while comprehensive ablation studies further support the contributions of proxy construction and multi-regime learning. By accounting for observation-induced feature variations during training, VoxSAOR offers a practical approach to improving the robustness and reliability of 3D scene perception for autonomous driving and embodied perception.

Future work will explore extending the robustification paradigm to spatiotemporal 3D scene representations. We will also investigate its potential benefits for downstream embodied tasks, such as navigation and motion planning, thereby bridging robust
3D perception and autonomous decision-making.

\bibliographystyle{IEEEtran}
\bibliography{bib}

\end{document}